# Neurogenesis-Inspired Dictionary Learning: Online Model Adaption in a Changing World

**Sahil Garg**
The Department of Computer Science, University of Southern California, Los Angeles, CA USA
sahilgar@usc.edu

**Irina Rish, Guillermo Cecchi, Aurelie Lozano**
IBM Thomas J. Watson Research Center, Yorktown Heights, NY USA
{rish, gcecchi, aclozano}@us.ibm.com

## Abstract

In this paper, we focus on online representation learning in non-stationary environments which may require continuous adaptation of model's architecture. We propose a novel online dictionary-learning (sparse-coding) framework which incorporates the addition and deletion of hidden units (dictionary elements), and is inspired by the *adult neurogenesis* phenomenon in the dentate gyrus of the hippocampus, known to be associated with improved cognitive function and adaptation to new environments. In the online learning setting, where new input instances arrive sequentially in batches, the "neuronal birth" is implemented by adding new units with random initial weights (random dictionary elements); the number of new units is determined by the current performance (representation error) of the dictionary, higher error causing an increase in the birth rate. "Neuronal death" is implemented by imposing $l_1/l_2$-regularization (group sparsity) on the dictionary within the block-coordinate descent optimization at each iteration of our online alternating minimization scheme, which iterates between the code and dictionary updates. Finally, hidden unit connectivity adaptation is facilitated by introducing sparsity in dictionary elements. Our empirical evaluation on several real-life datasets (images and language) as well as on synthetic data demonstrates that the proposed approach can considerably outperform the state-of-art fixed-size (non-adaptive) online sparse coding of Mairal et al. (2009) in the presence of non-stationary data. Moreover, we identify certain properties of the data (e.g., sparse inputs with nearly non-overlapping supports) and of the model (e.g., dictionary sparsity) associated with such improvements.

## 1 Introduction

The ability to adapt to a changing environment is essential for successful functioning in both natural and artificial intelligent systems. In human brains, adaptation is achieved via neuroplasticity, which takes different forms, including synaptic plasticity, i.e. changing connectivity strength among neurons, and neurogenesis, i.e. the birth and maturation of new neurons (accompanied with the death of some new or old neurons). Particularly, adult neurogenesis (Kempermann, 2006) (i.e., neurogenesis in the adult brain) in the dentate gyrus of the hippocampus is associated with improved cognitive functions such as pattern separation (Sahay et al., 2011), and is often implicated as a "candidate mechanism for the specific dynamic and flexible aspects of learning" (Stuchlik, 2014).

In the machine-learning context, synaptic plasticity is analogous to parameter tuning (e.g., learning neural net weights), while neurogenesis can be viewed as an online model selection via addition (and deletion) of hidden units in specific hidden-variable models used for representation learning (where hidden variables represent extracted features), from linear and nonlinear component analysis methods such as PCA, ICA, sparse coding (dictionary learning), nonlinear autoencoders, to deep neural nets and general hidden-factor probabilistic models. However, optimal model selection in large-scale hidden-variable models (e.g., adjusting the number of layers, hidden units, and their connectivity), is intractable due to enormous search space size. Growing a model gradually can be a more feasible alternative; after all, every real brain's "architecture" development process starts with a single cell. Furthermore, the process of adapting the model's architecture to dynamically changing environments is necessary for achieving a lifelong, continual learning. Finally, an online approach to dynamically expanding and contracting model's architecture can serve as a potentially more effective alternative to the standard off-line model selection (e.g., MDL-based off-line sparse coding (Ramirez & Sapiro, 2012)), as well as to the currently popular network compression (distillation) approaches (Hinton et al., 2015; Srivastava et al., 2014; Ba & Caruana, 2014; Bucilu et al., 2006), where a very large-scale architecture, such as a deep neural network with millions of parameters, must be first selected in ad-hoc ways and trained on large amounts of data, only to be compressed later to a more compact and simpler model with similarly good performance; we hypothesize that adaptive growth and reduction of the network architecture is a viable alternative to the distillation approach, although developing such an alternative remains the topic of further research.

In this paper, we focus on dictionary learning, a.k.a. sparse coding (Olshausen & Field, 1997; Kreutz-Delgado et al., 2003; Aharon et al., 2006; Lee et al., 2006) – a representation learning approach which finds a set of basis vectors (atoms, or dictionary elements) and representations (encodings) of the input samples as sparse linear combinations of those elements[1]. More specifically, our approach builds upon the computationally efficient *online dictionary-learning* method of Mairal et al. (2009), where the data samples are processed sequentially, one at a time (or in small batches). Online approaches are particularly important in large-scale applications with millions of potential training samples, where off-line learning can be infeasible; furthermore, online approaches are a natural choice for building systems capable of continual, lifelong learning.

Herein, we propose a novel online dictionary learning approach inspired by adult neurogenesis, which extends the state-of-art method of Mairal et al. (2009) to nonstationary environments by incorporating online model adaption, i.e. the addition and deletion of dictionary elements (i.e., hidden units) in response to the dynamically changing properties of the input data[2]. More specifically, at each iteration of online learning (i.e., for every batch of data samples), we add a group of random dictionary elements (modeling neuronal birth), where the group size depends on the current representation error, i.e. the mismatch between the new input samples and their approximation based on the current dictionary: higher error triggers more neurogenesis. The neuronal death, which involves removing "useless" dictionary elements, is implemented as an $l_1/l_2$ group-sparsity regularization; this step is essential in neurogenesis-inspired learning, since it reduces a potentially uncontrolled growth of the dictionary, and helps to avoid overfitting (note that neuronal death is also a natural part of the adult neurogensis process, where neuronal survival depends on multiple factors, including the complexity of a learning environment (Kempermann, 2006)). Moreover, we introduce sparsity in dictionary elements, which reflects sparse connectivity between hidden units/neurons and their inputs; this is a more biologically plausible assumption than the fully-connected architecture of standard dictionary learning, and it also works better in our experiments. Thus, adaptation in our model involves not only the addition/deletion of the elements, but adapting their connectivity as well.

We demonstrate on both simulated data and on two real-life datasets (natural images and language processing) that, in presence of a non-stationary input, our approach can significantly outperform non-adaptive, fixed-dictionary-size online method of Mairal et al. (2009). Moreover, we identify certain data properties and parameter settings associated with such improvements. Finally, we demonstrate that the novel approach not only improves the representation accuracy, but also can boost the classification accuracy based on the extracted features.

Note that, although the group-sparsity constraint enforcing deletion of some dictionary elements was introduced earlier in the group-sparse coding method of Bengio et al. (2009), it was only implemented and tested in the off-line rather than online setting, and, most importantly, it was not ac-companied by the neurogenesis. On the other hand, while some prior work considered online node addition in hidden-variable models, and specifically, in neural networks, from cascade correlations (Fahlman & Lebiere, 1989) to the recent work by Draelos et al. (2016a;b), no model pruning was incorporated in those approaches in order to balance the model expansion. Overall, we are not aware of any prior work which would propose and systematically evaluate, empirically and theoretically, a dynamic process involving both addition and deletion of hidden units in the online model selection setting, either in sparse coding or in a neural network setting.

[1] Note that the corresponding neural network interpretation of sparse coding framework is a (single-hidden-layer) linear autoencoder with sparsity constraints: the hidden units are associated with dictionary elements, each element represented by a weight vector associated with unit's outgoing links in the output layer, and the sparse vector of hidden unit activations corresponding to the encoding of an input.

[2] An early version of our neurogenetic online dictionary learning approach was presented as a poster at the 2011 Society for Neuroscience meeting (Rish et al., 2011), although it did not appear before as a peer-reviewed publication.

To summarize, the main contributions of this paper are as follows:

- we propose a *novel online model-selection approach to dictionary learning*[3], inspired by the *adult neurogenesis* phenomenon; our method *significantly outperforms the state-of-art baseline*, especially in *non-stationary* settings;
- we perform an *extensive empirical evaluation, on both synthetic and real data*, in order to identify the conditions when the proposed adaptive approach is most beneficial, both for data reconstruction and for classification based on extracted features; we conclude that these conditions include a combination of *sparse dictionary elements (and thus a more biologically plausible sparse network connectivity as opposed to fully connected units), accompanied by sufficiently dense codes*;
- furthermore, we provide an intuitive discussion, as well as *theoretical analysis of certain combinations of the input data properties and the algorithm's parameters* when the proposed approach is most beneficial;
- from the neuroscientific perspective, we propose a computational model which supports earlier empirical observations indicating that adult neurogenesis is particularly beneficial in changing environments, and that certain amount of neuronal death, which accompanies the neuronal birth, is an important component of an efficient neurogenesis process;
- overall, to the best of our knowledge, *we are the first to perform an in-depth evaluation of the interplay between the birth and death of hidden units in the context of online model selection in representation learning, and, more specifically, in online dictionary learning.*

This paper is organized as follows. In Sec. 2, we summarize the state-of-art non-adaptive (fixed-size) online dictionary learning method of Mairal et al. (2009). Thereafter, in Sec. 3, we describe our adaptive online dictionary learning algorithm. In Sec. 4, we present our empirical results on both synthetic and real datasets, including images and language data. Next, in Sec. 5, we provide some theoretical, as well as an intuitive analysis of settings which can benefit most from our approach. Finally, we conclude with a summary of our contributions in Sec. 6. The implementation details of the algorithms and additional experimental results are described in the Appendix.

## 2 Background on Dictionary Learning

Traditional off-line dictionary learning (Olshausen & Field, 1997; Aharon et al., 2006; Lee et al., 2006) aims at finding a *dictionary* $\boldsymbol{D} \in \mathbb{R}^{m\times k}$, which allows for an accurate representation of a training data set $\boldsymbol{X} = \{\boldsymbol{x}_1, \cdots, \boldsymbol{x}_n \in \mathbb{R}^m\}$, where each sample $\boldsymbol{x}_i$ is approximated by a linear combination $\boldsymbol{x}_i \approx \boldsymbol{D}\boldsymbol{\alpha}_i$ of the columns of $\boldsymbol{D}$, called *dictionary elements* $\{\boldsymbol{d}_1, \cdots, \boldsymbol{d}_k \in \mathbb{R}^m\}$. Here $\boldsymbol{\alpha}_i$ is the *encoding* (*code vector*, or simply *code*) of $\boldsymbol{x}_i$ in the dictionary. Dictionary learning is also referred to as *sparse coding*, since it is assumed that the code vectors are *sparse*, i.e. have a relatively small number of nonzeros; the problem is formulated as minimizing the objective

$$f_n(D) = \frac{1}{n}\sum_{i=1}^{n}\frac{1}{2}||\boldsymbol{x}_i - \boldsymbol{D}\boldsymbol{\alpha}_i||_2^2 + \lambda_c||\boldsymbol{\alpha}_i||_1 \quad (1)$$

where the first term is the mean square error loss incurred due to approximating the input samples by their representations in the dictionary, and the second term is the $l_1$-regularization which enforces the codes to be sparse. The joint minimization of $f_n(D)$ with respect to the dictionary and codes is non-convex; thus, a common approach is alternating minimization involving convex subproblems of finding optimal codes while fixing a dictionary, and vice versa.

[3]The Matlab code is available at `https://github.com/sgarg87/neurogenesis_inspired_dictionary_learning`.

However, the classical dictionary learning does not scale to very large datasets; moreover, it is not immediately applicable to online learning from a continuous stream of data. The *online dictionary learning (ODL)* method proposed by Mairal et al. (2009) overcomes both of these limitations, and serves as a basis for our proposed approach, presented in Alg. 1 in the next section. While the highlighted lines in Alg. 1 represent our extension of ODL , the non-highlighted ones are common to both approaches, and are discussed first. The algorithms start with some dictionary $\boldsymbol{D}^0$, e.g. a randomly initialized one (other approaches include using some of the inputs as dictionary elements (Mairal et al., 2010; Bengio et al., 2009)). At each iteration $t$, both online approaches consider the next input sample $\boldsymbol{x}_t$ (more generally, a batch of samples) as in the step 3 of Alg. 1 and compute its sparse code $\boldsymbol{\alpha}_t$ by solving the LASSO (Tibshirani, 1996) problem (the step 4 in Alg. 1), with respect to the current dictionary. In Alg. 1, we simply use $\boldsymbol{D}$ instead of $\boldsymbol{D}^{(t)}$ to simplify the notation. Next, the standard ODL algorithm computes the dictionary update, $\boldsymbol{D}^{(t)}$, by optimizing the *surrogate* objective function $\hat{f}_t(D)$ which is defined just as the original objective in eq. (1), for $n = t$, but with one important difference: unlike the original objective, where each code $\boldsymbol{\alpha}_i$ for sample $\boldsymbol{x}_i$ is computed with respect to the *same* dictionary $\boldsymbol{D}$, the surrogate function includes the codes $\boldsymbol{\alpha}_1, \boldsymbol{\alpha}_2, \cdots, \boldsymbol{\alpha}_t$ computed at the *previous* iterations, using the dictionaries $\boldsymbol{D}^{(0)}, ..., \boldsymbol{D}^{(t-1)}$, respectively; in other words, it *does not recompute* the codes for previously seen samples after each dictionary update. This speeds up the learning without worsening the (asymptotic) performance, since the surrogate objective converges to the original one in (1), under certain assumptions, including data stationarity (Mairal et al., 2009). Note that, in order to prevent the dictionary entries from growing arbitrarily large, Mairal et al. (2009; 2010) impose the norm constraint, i.e. keep the columns of $\boldsymbol{D}$ within the convex set $\mathcal{C} = \{\boldsymbol{D} \in \mathbb{R}^{m\times k} \quad s.t. \ \forall j \ \boldsymbol{d}_j^T \boldsymbol{u}_j \leq 1\}$. Then the dictionary update step computes $\boldsymbol{D}^{(t)} = \arg\min_{\boldsymbol{D}\in\mathcal{C}} \hat{f}_t(D)$, ignoring $l_1$-regularizer over the code which is fixed at this step, as

$$\arg\min_{\boldsymbol{D}\in\mathcal{C}} \frac{1}{t}\sum_{i=1}^{t}\frac{1}{2}||\boldsymbol{x}_i - \boldsymbol{D}\boldsymbol{\alpha}_i||_2^2 = \arg\min_{\boldsymbol{D}\in\mathcal{C}} \frac{1}{2}Tr(\boldsymbol{D}^T\boldsymbol{D}\boldsymbol{A}) - Tr(\boldsymbol{D}^T\boldsymbol{B}), \tag{2}$$

where $\boldsymbol{A} = \sum_{i=1}^t \boldsymbol{\alpha}_i\boldsymbol{\alpha}_i^T$ and $\boldsymbol{B} = \sum_{i=1}^t \boldsymbol{x}_i\boldsymbol{\alpha}_i^T$ are the "bookkeeping" matrices (we also call them "memories" of the model), compactly representing the input samples and encoding history. At each iteration, once the new input sample $\boldsymbol{x}_i$ is encoded, the matrices are updated as $\boldsymbol{A} \leftarrow \boldsymbol{A} + \boldsymbol{\alpha}_t\boldsymbol{\alpha}_t^T$ and $\boldsymbol{B} \leftarrow \boldsymbol{B} + \boldsymbol{x}_t\boldsymbol{\alpha}_t^T$ (see the step 11 of Alg. 1). In (Mairal et al., 2009; 2010), a *block coordinate descent* is used to optimize the convex objective in eq. 2; it iterates over the dictionary elements in a fixed sequence, optimizing each while keeping the others fixed as shown in eq. (3) (essentially, the steps 14 and 17 in Alg. 1; the only difference is that our approach will transform $\boldsymbol{u}_j$ into $\boldsymbol{w}_j$ in order to impose additional regularizer before computing step 17), until convergence.

$$\boldsymbol{u}_j \leftarrow \frac{\boldsymbol{b}_j - \sum_{k\neq j}\boldsymbol{d}_k a_{jk}}{a_{jj}}; \quad \boldsymbol{d}_j \leftarrow \frac{\boldsymbol{u}_j}{\max(1, ||\boldsymbol{u}_j||_2)} \tag{3}$$

Herein, when the off-diagonal entries $\boldsymbol{a}_{jk}$ in $\boldsymbol{A}$ are as large as the diagonal $a_{jj}$, the dictionary elements get "tied" to each other, playing complementary roles in the dictionary, thereby constraining the updates of each other.

*It is important to note that, for the experiment settings where we consider dictionary elements to be sparse in our algorithm NODL (discussed next in Sec. 3), we will actually use as a baseline algorithm a modified version of the fixed-size ODL, which allows for sparse dictionary elements, i.e. includes the sparsification step 15 in Alg. 1, thus optimizing the following objective in dictionary update step instead of the one in eq. (2):*

$$\arg\min_{\boldsymbol{D}\in\mathcal{C}} \frac{1}{t}\sum_{i=1}^{t}\frac{1}{2}||\boldsymbol{x}_i - \boldsymbol{D}\boldsymbol{\alpha}_i||_2^2 + \sum_j \lambda_j||\boldsymbol{d}_j||_1. \tag{4}$$

From now on, ODL will refer to the above extended version of the fixed-size method of Mairal et al. (2009) wherever we have sparsity in dictionary elements (otherwise, the standard method of Mairal et al. (2009) is the baseline); in our experiments, dictionary sparsity of both the baseline and the proposed method (discussed in the next section) will be matched. Note that Mairal et al. (2010) mention that the convergence guaranties for ODL hold even with the sparsity constraints on dictionary elements.

## 3 OUR APPROACH: NEUROGENIC ONLINE DICTIONARY LEARNING (NODL)

Our objective is to extend the state-of-art online dictionary learning, designed for stationary input distributions, to a more adaptive framework capable of handling nonstationary data effectively, and learning to represent new types of data without forgetting how to represent the old ones. Towards this end, we propose a novel algorithm, called Neurogenetic Online Dictionary Learning (see Alg. 1), which can flexibly extend and reduce a dictionary in response to the changes in an input distribution, and possibly to the inherent representation complexity of the data. The main changes, as compared to the non-adaptive, fixed-dictionary-size algorithm of Mairal et al. (2009), are highlighted in Alg. 1; the two parts involve (1) neurogenesis, i.e. the addition of dictionary elements (hidden units, or "neurons") and (2) the death of old and/or new elements which are "less useful" than other elements for the task of data reconstruction.

At each iteration in Alg. 1, the next batch of samples is received and the corresponding codes, in the dictionary, are computed; next, we add $k_n$ new dictionary elements sampled at random from $\mathbb{R}^m$ (i.e., $k_n$ random linear projections of the input sample). The choice of the parameter $k_n$ is important; one approach is to tune it (e.g., by cross-validation), while another is to adjust it dynamically, based on the dictionary performance: e.g., if the environment is changing, the old dictionary may not be able to represent the new input well, leading to decline in the representation accuracy, which triggers neurogenesis. Herein, we use as the performance measure the Pearson correlation between a new sample and its representation in the current dictionary $r(\boldsymbol{x}_t, \boldsymbol{D}^{(t-1)}\boldsymbol{\alpha}_t)$, i.e. denoted as $p_c(\boldsymbol{x}_t, \boldsymbol{D}^{(t-1)}, \boldsymbol{\alpha}_t)$ (for a batch of data, the average over $p_c(.)$ is taken). If it drops below a certain pre-specified threshold $\gamma$ (where $0 \ll \gamma \leq 1$), the neurogenesis is triggered (the step 5 in Alg. 1). The number $k_n$ of new dictionary elements is proportional to the error $1 - p_c(\cdot)$, so that worse performance will trigger more neurogenesis, and vice versa; the maximum number of new elements is bounded by $c_k$ (the step 6 in Alg. 1). We refer to this approach as *conditional neurogenesis* as it involves the *conditional birth* of new elements. Next, $k_n$ random elements are generated and added to the current dictionary (the step 7), and the memory matrices $\boldsymbol{A}, \boldsymbol{B}$ are updated, respectively, to account for larger dictionary (the step 8). Finally, the sparse code is recomputed for $\boldsymbol{x}_t$ (or, all the samples in the current batch) with respect to the extended dictionary (the step 9).

The next step is the dictionary update, which uses, similarly to the standard online dictionary learning, the block-coordinate descent approach. However, the objective function includes additional regularization terms, as compared to (2):

$$\boldsymbol{D}^{(t)} = \arg\min_{\boldsymbol{D}\in\mathcal{C}} \frac{1}{t}\sum_{i=1}^{t}\frac{1}{2}||\boldsymbol{x}_i - \boldsymbol{D}\boldsymbol{\alpha}_i||_2^2 + \lambda_g\sum_j ||\boldsymbol{d}_j||_2 + \sum_j \lambda_j||\boldsymbol{d}_j||_1. \tag{5}$$

The first term is the standard reconstruction error, as before. The second term, $l_1/l_2$-regularization, promotes group sparsity over the dictionary entries, where each group corresponds to a column, i.e. a dictionary element. The group-sparsity (Yuan & Lin, 2006) regularizer causes some columns in $\boldsymbol{D}$ to be set to zero (i.e. the columns less useful for accurate data representation), thus effectively eliminating the corresponding dictionary elements from the dictionary ("killing" the corresponding hidden units). As it was mentioned previously, Bengio et al. (2009) used the $l_1/l_2$-regularizer in dictionary learning, though not in online setting, and without neurogenesis.

Finally, the third term imposes $l_1$-regularization on dictionary elements thus promoting sparse dictionary, besides the sparse coding. Introducing sparsity in dictionary elements, corresponding to the sparse connectivity of hidden units in the neural net representation of a dictionary, is motivated by both their biological plausibility (neuronal connectivity tends to be rather sparse in multiple brain networks), and by the computational advantages this extra regularization can provide, as we observe later in experiments section (Sec. 4).

As in the original algorithm of Mairal et al. (2009), the above objective is optimized by the block-coordinate descent, where each block of variables corresponds to a dictionary element, i.e., a column in $\boldsymbol{D}$; the loop in steps 12-19 of the Alg. 1 iterates until convergence, defined by the magnitude of change between the two successive versions of the dictionary falling below some threshold. For each column update, the first and the last steps (the steps 14 and 17) are the same as in the original method of Mairal et al. (2009), while the two intermediate steps (the steps 15 and 16) are implementing additional regularization. Both steps 15 and 16 (sparsity and group sparsity regularization) are

**Algorithm 1** Neurogenetic Online Dictionary Learning (NODL)

**Require:** Data stream $\boldsymbol{x}_1, \boldsymbol{x}_2, \cdots, \boldsymbol{x}_n \in \mathbb{R}^m$; initial dictionary $\boldsymbol{D} \in \mathbb{R}^{m \times k}$; conditional neurogenesis threshold, $\gamma$; max number of new elements added per data batch, $c_k$; group sparsity regularization parameter, $\lambda_g$; number of non-zeros in a dictionary element, $\beta_d$; number of non-zeros in a code, $\beta_c$.

Initialize: $\boldsymbol{A} \leftarrow 0, \boldsymbol{B} \leftarrow 0$ `% reset the ''memory''`
`% assuming single data in a batch, for the simpler exposition`
**for** $t = 1$ to $n$ **do**
  Input $\boldsymbol{x}_t$ `% representing the` $t_{th}$ `batch of data`
  `% Sparse coding of data:`
  $\boldsymbol{\alpha}_t = \arg_{\boldsymbol{\alpha} \in \mathbb{R}^k} \min \frac{1}{2}||\boldsymbol{x}_t - \boldsymbol{D}\boldsymbol{\alpha}||_2^2 + \lambda_c||\boldsymbol{\alpha}||_1$ `%` $\lambda_c$ `tuned to have` $\beta_c$ `non-zeros in` $\alpha_t$
  `% Conditional neurogenesis: if accuracy below threshold, add more elements (should not be more than the number of data in a batch`
  **if** $p_c(\boldsymbol{x}_t, \boldsymbol{D}, \boldsymbol{\alpha}_t) \le \gamma$ **then**
    $k_n = (1 - p_c(\boldsymbol{x}_t, \boldsymbol{D}, \boldsymbol{\alpha}_t))c_k$ `% the count of the births of neurons`
    $\boldsymbol{D}_n \leftarrow initializeRand(k_n)$,
    $\boldsymbol{D} \leftarrow [\boldsymbol{D} \quad \boldsymbol{D}_n]$
    $\boldsymbol{A} \leftarrow \begin{bmatrix} \boldsymbol{A} & \boldsymbol{0} \\ \boldsymbol{0} & \boldsymbol{0} \end{bmatrix}$, $\boldsymbol{B} \leftarrow [\boldsymbol{B} \quad \boldsymbol{0}], k \leftarrow k + k_n$
    `% Repeat sparse coding, now including the new dictionary elements`
    $\boldsymbol{\alpha}_t = \arg_{\boldsymbol{\alpha} \in \mathbb{R}^k} \min \frac{1}{2}||\boldsymbol{x}_t - \boldsymbol{D}\boldsymbol{\alpha}||_2^2 + \lambda_c||\boldsymbol{\alpha}||_1$
  **end if** `% End of neurogenesis`
  `% ''Memory'' update:`
  $\boldsymbol{A} \leftarrow \boldsymbol{A} + \boldsymbol{\alpha}_t\boldsymbol{\alpha}_t^T,\ \boldsymbol{B} \leftarrow \boldsymbol{B} + \boldsymbol{x}_t\boldsymbol{\alpha}_t^T$
  `% Dictionary update by block-coordinate descent with` $l_1/l_2$ `group sparsity`
  **repeat**
    **for** $j = 1$ to $k$ **do**
      $\boldsymbol{u}_j \leftarrow \frac{\boldsymbol{b}_j - \sum_{k \neq j} \boldsymbol{d}_k a_{jk}}{a_{jj}}$
      `% Sparsifying elements (optional):`
      $\boldsymbol{v}_j \leftarrow Prox_{\lambda_j||.||_1}(\boldsymbol{u}_j) = sgn(\boldsymbol{u}_j)(|\boldsymbol{u}_j| - \lambda_j)_+$, `%` $\lambda_j$ `tuned to get` $\beta_d$ `non-zeros in` $\boldsymbol{v}_j$
      `% Killing useless elements with` $l_1/l_2$ `group sparsity`
      $\boldsymbol{w}_j \leftarrow \boldsymbol{v}_j \left(1 - \frac{\lambda_g}{||\boldsymbol{v}_j||_2}\right)_+$
      $\boldsymbol{d}_j \leftarrow \frac{\boldsymbol{w}_j}{\max(1, ||\boldsymbol{w}_j||_2)}$
    **end for**
  **until convergence**
**end for**
**return** $\boldsymbol{D}$

implemented using the standard proximal operators as described in Jenatton et al. (2011). Note that we actually use as input the desired number of non-zeros, and determine the corresponding sparsity parameter $\lambda_c$ and $\lambda_j$ using a binary search procedure (see Appendix).

Overall, *the key features of our algorithm is the interplay of both the (conditional) birth and (group-sparsity) death of dictionary elements in an online setting.*

### 3.1 Discussion of Important Algorithmic Details

**A rationale behind sparsity of dictionary elements.** We focus here on sparse dictionary elements, which, in the network terms, correspond to sparse connectivity between hidden units and their inputs; one reason for this choice was that sparse connectivity appears to be a more biologically plausible assumption than a fully-connected architecture implied by dense dictionary, in many brain areas, and specifically between dentate gyrus and CA3. The other reason relates to computational advantages.

Note that (Mairal et al., 2009) state that convergence guaranties for the original ODL algorithm would also hold for the case of sparse dictionary elements. However, no empirical evaluation is provided for this case; furthermore, we are not aware of any previous work on sparse coding which would involve and extensive empirical evaluation for such setting. Prior focus on dense rather than sparse dictionary elements is perhaps more natural when the input consists of a large number of relatively small image patches, and thus each element also represents a small patch. In our work, however, dictionary is being learned on full images, and thus a nonzero pattern in a sparse dictionary element corresponds to a small patch within a larger image, with multiple sparse elements (patches) covering the image. Thus, rather than explicitly representing an image as a set of patches and then learning a dictionary of dense elements for accurate representation of such patches, a dictionary of full-image-size, but sparse dictionary elements can be used to implicitly represents an image as a linear combination of those elements, with possible overlap of non-zero pixels between elements; the non-zero pixels in a sparse element of a dictionary are learned automatically. Computational advantages of using sparse dictionaries are demonstrated in our experiment results (Sec. 4), where classifiers learned on top of representations extracted with sparse dictionaries yield smaller errors.

**The memory matrix $\boldsymbol{A}$ and its properties.** The matrix $\boldsymbol{A}$ keeps the "memory" of the encodings $\alpha_t$ for the previous data samples, in a sense, as it accumulates the sum of $\alpha_t \alpha_t^T$ matrices from each iteration $t$. It turns out that the matrix $\boldsymbol{A}$ can have a significant effect on dictionary learning in both ODL and NODL algorithms. As it is pointed out in (Mairal et al., 2009), the quadratic surrogate function in (2) is strictly convex with a lower-bounded Hessian $\boldsymbol{A}$ ensuring convergence to a solution. From the practical standpoint, when the matrix $\boldsymbol{A}$ has a high condition number (the ratio of the largest to smallest singular value in the singular value decomposition of a matrix), despite its lower-bounded eigenvalues, the adaptation of a dictionary elements using the standard ODL algorithm can be difficult, as we see in our experiments. Specifically, when the dictionary elements are sparse, this effect is more pronounced, since the condition number of $\boldsymbol{A}$ becomes high due to the complementary roles of sparse dictionary elements in the reconstruction process (see the comparison of $\boldsymbol{A}$ from dense elements and sparse elements in 6(a) and 6(b), respectively). In such scenarios, the submatrix of $\boldsymbol{A}$ corresponding to the new elements in a dictionary, added by our NODL algorithm, can have a better condition number, leading to an improved adaptation of the dictionary.

**Code Sparsity.** Code sparsity is controlled by the parameter $\beta_c$, the number of nonzeros, which determines the corresponding regularization weight $\lambda_c$ in step 4 of Alg. 1; note that $\lambda_c$ is determined via binary search for each input sample separately, as shown in Algorithm 2, and thus may vary slightly for different instances given a fixed $\beta_c$.

Selecting an appropriate level of code sparsity depends on the choice of other parameters, such as the input batch size, sparsity of the dictionary elements, the extent of non-stationarity and complexity of the data, and so on. When the dictionary elements are themselves sparse, denser codes may be more appropriate, since each sparse dictionary element represents only a relatively small subset of image pixels, and thus a large number of those subsets covering the whole image may be needed for an accurate input representation.

Interestingly, using very sparse codes in combination with non-sparse dictionary elements in the standard ODL approach can sometimes lead to creation of "dead" (zero $l_2$-norm) elements in the dictionary, especially if the input batch size is small. This is avoided by our NODL algorithm, since such dead elements are implicitly removed via group sparsity at the dictionary update step, along with the "weak" (very small $l_2$-norm) elements. Also, a very high code sparsity in combination with dense dictionary elements can lead to a significant decrease in the reconstruction accuracy for both ODL and our NODL when the online data stream is non-stationary. Such shortcomings were not encountered in (Mairal et al., 2009; 2010), where only stationary data streams were studied, both in theoretical and empirical results. On the other hand, high sparsity in dictionary elements does not seem to cause a degradation in the reconstruction accuracy, as long as the codes are not too sparse.

**The choice and tuning of metric for conditional neuronal birth.** In the "conditional birth" approach described above, the number of new elements $k_n$ is determined based on the performance of the current dictionary, using the Pearson correlation between the actual and reconstructed data, for the current batch. This is, of course, just one particular approach to measuring data nonstationarity and the need for adaptation, but we consider it a reasonable heuristic. Low reconstruction error indicates that the old dictionary is still capable of representing the new data, and thus less adaptation might be needed, while a high error indicates that the data distribution might have changed, and trigger neurogenesis in order to better adapt to a new environment. We choose the Pearson correlation as the measure of reconstruction accuracy since its value is easily interpretable, is always in the range $[0, 1]$ (unlike, for example, the mean-square error), which simplifies tuning the threshold parameter $\gamma$. Clearly, one can also try other interpretable metrics, such as, for example, the Spearman correlation.

**Tuning parameters: group sparsity $\lambda_g$ and others.** The group sparsity regularization parameter $\lambda_g$ controls the amount of removal ("death") of elements in NODL : in step 16 of the Alg. 1, all elements with $l_2$-norm below $\lambda_g$ (i.e., "weak" elements), are set to zero ("killed"). Since the dictionary elements are normalized to have $l_2$-norm less than one, we only need to consider $\lambda_g \in [0, 1]$. (Note that the step of killing dictionary elements precedes the normalization step in the algorithm. Thus, the tuning of $\lambda_g$ is affected by the normalization of the elements from the previous iteration.) Note that increasing the sparsity of the dictionary elelments, i.e. decreasing $\beta_d$ (the number of nozeros in dictionary elements) may require the corresponding reduction of $\lambda_g$, while an increase in the input dimensionality $m$ may also require an increase in the $\lambda_g$ parameter. Tuning the rest of the parameters is relatively easy. Clearly, the batch size should be kept relatively small, and, ideally, not exceed the "window of stationarity" size in the data (however, the frequency of the input distribution change may need to be also estimated from the data, and thus the batch size may need to be tuned adaptively, which is outside of the scope of this paper). Mairal et al. (2009) suggest to use a batch size of 256 in their experiments while getting similar performance with values 128 and 512. As to the maximum number of new elements $c_k$ added at each iteration, it is reasonable to keep it smaller than the batch size.

## 4 Experiments

We now evaluate empirically the proposed approach, NODL, against ODL, the standard (non-adaptive) online dictionary learning of Mairal et al. (2009). Moreover, in order to evaluate separately the effects of either only adding, or only deleting dictionary elements, we also evaluate two restricted versions of our method: NODL+ involves only addition but no deletion (equivalent to NODL with no group-sparsity, i.e. $\lambda_g = 0$), and NODL- which, vice versa, involves deletion only but no addition (equivalent to NODL with the number of new elements $c_k = 0$). The above algorithms are evaluated in a non-stationary setting, where a sequence of training samples from one environment (first domain) is followed by another sequence from a different environment (second domain), in order to test their ability to adapt to new environments without "forgetting" the previous ones.

### 4.1 Real-life Images

Our first domain includes the images of Oxford buildings [4] (urban environment), while the second uses a combination of images from Flowers [5] and Animals [6] image databases (natural environment); examples of both types of images are shown in Fig. 1(a) and 1(b). We converted the original color images into black&white format and compressed them to smaller sizes, 32x32 and 100x100. Note that, unlike (Mairal et al., 2009), we used full images rather than image patches as our inputs.

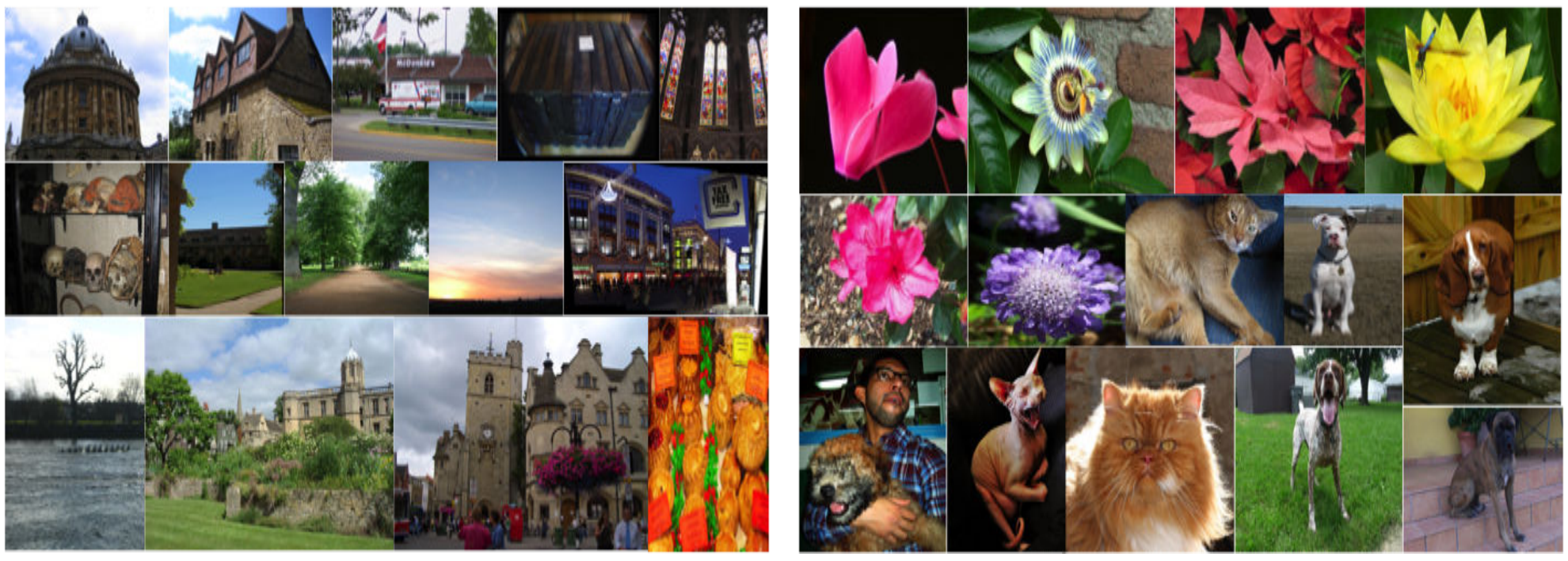

(a) Urban: Oxford Buildings (b) Nature: Flowers and Animals

Figure 1: The image data sets for the evaluation of the online dictionary learning algorithms.

We selected 5700 images for training and another 5700 for testing; each subset contained 1900 images of each type (i.e., Oxford, Flowers, Animals). In the training phase, as mentioned above,

[4] http://www.robots.ox.ac.uk/~vgg/data/oxbuildings/index.html

[5] http://www.robots.ox.ac.uk/~vgg/data/flowers/102/

[6] http://www.robots.ox.ac.uk/~vgg/data/pets/

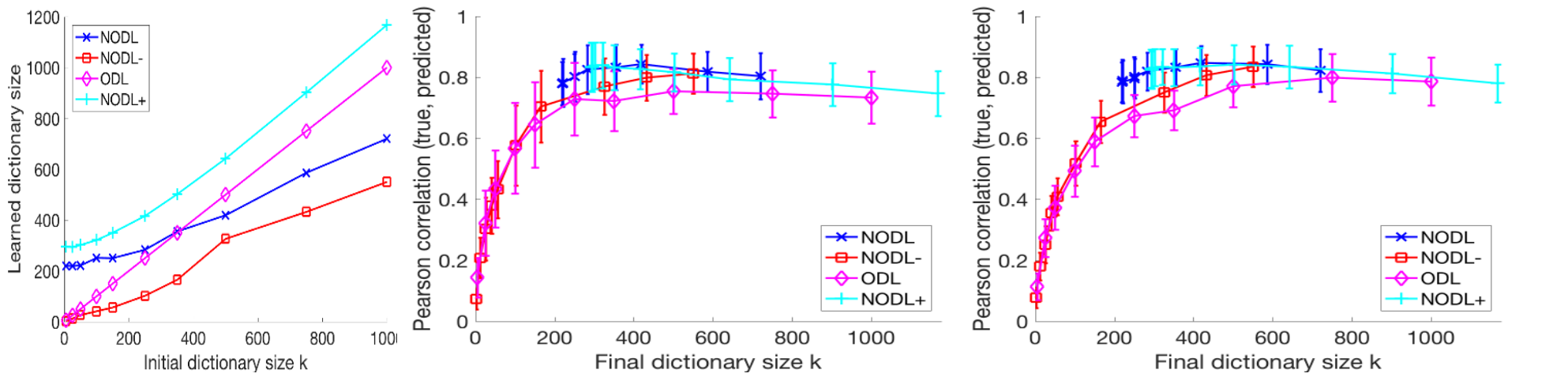

(a) Learned Dictionary Size (b) 1st domain (Oxford) (c) 2nd domain (Flowers)

Figure 2: Reconstruction accuracy of NODL and ODL on 32x32 images (sparse dictionary).

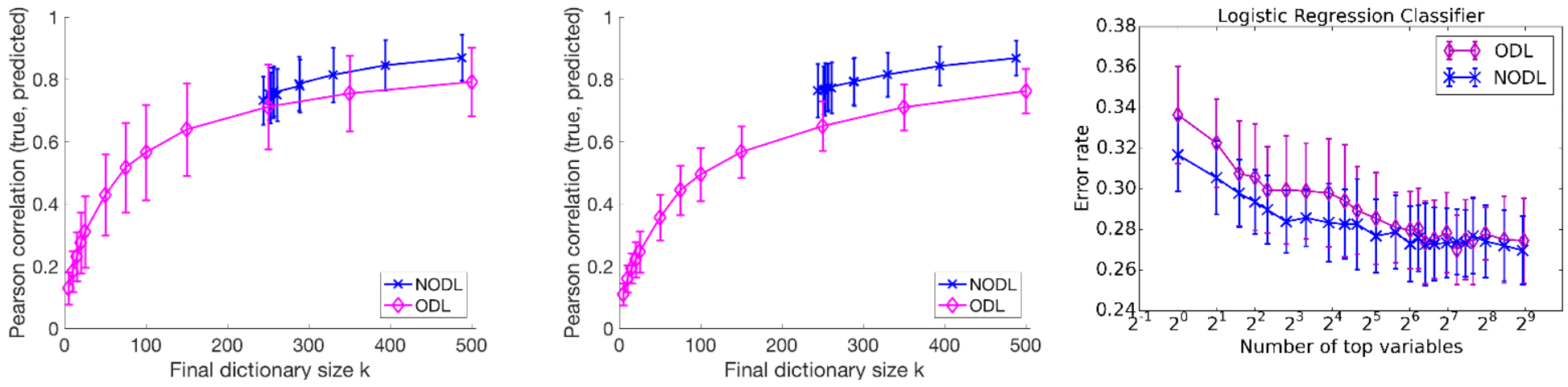

(a) 1st domain (Oxford) (b) 2nd domain (Flowers) (c) Classification Error

Figure 3: Reconstruction accuracy of NODL and ODL on 100x100 images with sparse dictionary elements (50 non-zeros) and non-sparse codes.

each online dictionary learning algorithm receives a sequence of 1900 samples from the first, urban domain (Oxford), and then a sequence of 3800 samples from the second, natural domain (1900 Flowers and 1900 Animals, permuted randomly). At each iteration, a batch of 200 images is received as an input. (For comparison, Mairal et al. (2009) used a batch of size 256, though image patches rather than full images.) The following parameters are used by our algorithm: Pearson correlation threshold $\gamma = 0.9$, group sparsity parameter $\lambda_g = 0.03$ and $\lambda_g = 0.07$, for 32x32 and 100x100 images, respectively. The upper bound on the number of new dictionary elements at each iteration is $c_k = 50$. (We observed that the results are only mildly sensitive to the specified parameter values.)

Once the training phase is completed, the resulting dictionary is evaluated on test images from both the first (urban) and the second (natural) domains; for the second domain, separate evaluation is performed for flowers and animals. First, we evaluate the *reconstruction* ability of the resulting dictionary $\boldsymbol{D}$, comparing the actual inputs $\boldsymbol{x}$ versus approximations $\boldsymbol{x}^* = \boldsymbol{D\alpha}$, using the mean square error (MSE), Pearson correlation, and the Spearman correlation. We present the results for Pearson correlations between the actual and reconstructed inputs, since all the three metrics show consistent patterns (for completeness, MSE results are shown in Appendix). Moreover, we evaluate the dictionaries in a *binary classification* setting (e.g., flowers vs animals), using as features the codes of test samples in a given dictionary. Finally, we explored a wide range of sparsity parameters for both the codes and the dictionary elements.

Our key observations are that: *(1) the proposed method frequently often outperforms (or is at least as good as) its competitors, on both the new data (adaptation) and the old ones (memory); (2) it is most beneficial when dictionary elements are sparse; (3) vice versa, when dictionary elements are dense, neurogenetic approach matches the baseline, fixed-size dictionary learning.* We now discuss the results in detail.

**Sparse Dictionary Elements**

In Fig. 2, we present the results for sparse dictionaries, where each column (an element in the dictionary) has 5 nonzeros out of the 1024 dimensions; the codes are relatively dense, with at most 200 nonzeros out of $k$ (the number of dictionary elements), and $k$ ranging from 5 to 1000 (i.e. the codes are not sparse for $k \leq 200$). Due to space limitations, we put in the Appendix (Sec. B.2) our results on a wider range of values for the dictionary and code sparsity (Fig. 12). In Fig. 2(a), we compare the dictionary size for different methods: the final dictionary size after completing the training phase (y-axis) is plotted against the initial dictionary size (x-axis). Obviously, the baseline (fixed-size) ODL method (magenta plot) keeps the size constant, deletion-only NODL- approach reduces the initial size (red plot), and addition-only NODL+ increases the size (light-blue plot).

However, the interplay between the addition and deletion in our NODL method (dark-blue) produces a more interesting behavior: it tends to adjust the representation complexity towards certain balanced range, i.e. very small initial dictionaries are expanded, while very large ones are, vice versa, reduced.

Our main results demonstrating the advantages of the proposed NODL method are shown next in Fig. 2(b) and Fig. 2(c), for the "old" (Oxford) and "new" (Flowers) environment (domain), respectively. (Very similar result are shown for Animals as well, in the Appendix). The x-axis shows the final dictionary size, and the y-axis is the reconstruction accuracy achieved by the trained dictionary on the test samples, measured by Pearson correlation between the actual and reconstructed data. NODL clearly outperforms the fixed-size ODL, especially on smaller dictionary sizes; remarkably, this happens on both domains, i.e. besides improved adaptation to the new data, NODL is also better at preserving the "memories" of the old data, *without increasing the representation complexity, i.e. for the same dictionary size*.

Interestingly, just deletion would not suffice, as deletion-only version, NODL-, is inferior to our NODL method. On the other hand, addition-only, or NODL+, method is as accurate as NODL, but tends to increase the dictionary size too much. The interplay between the addition and deletion processes in our NODL seems to achieve the best of the two worlds, achieving superior performance while keeping the dictionary size under control, in a narrower range (400 to 650 elements), expanding, as necessary, small dictionaries, while compressing large ones[7].

We will now focus on comparing the two main methods, the baseline ODL and the proposed NODL method. The advantages of our approach become even more pronounced on larger input sizes, e.g. 100x100 images, in similar sparse-dictionary, dense-code settings. (We keep the dictionary elements at the same sparsity rate, 50 nonzeros out of 10,000 dimensions, and just use completely non-sparse codes). In Fig. 3(a) and Fig. 3(b), we see that NODL considerably outperforms ODL on both the first (Oxford) and the (part of the ) second domain (Flowers); the results for Animals are very similar and are given in the Appendix in Fig. 10. In Appendix Sec. B.6, Fig. 17 depicts examples of actual animal images and the corresponding reconstructions by the fixed-size ODL and our NODL methods (not included here due to space restrictions). A better reconstruction quality of our method can be observed (e.g., a more visible dog shape, more details such as dog's legs, as opposed to a collection clusters produced by the ODL methods note however that printer resolution may reduce the visible difference, and looking at the images in online version of this paper is recommended).

Moreover, NODL can be also beneficial in classification settings. Given a dictionary, i.e. a sparse linear autoencoder trained in an unsupervised setting, we use the codes (i.e., feature vectors) computed on the test data from the second domain (Animals and Flowers) and evaluate multiple classifiers learned on those features in order to discriminate between the two classes. In Fig. 3(c), we show the logistic regression results using 10-fold cross-validation; similar results for several other classifiers are presented in the Appendix, Fig. 10. Note that we also perform filter-based feature subset selection, using the features statistical significance as measured by its p-value as the ranking function, and selecting subsets of top $k$ features, increasing $k$ from 1 to the total number of features (the code length, i.e. the number of dictionary elements). The x-axis in Fig. 3(c) shows the value of $k$, while the y-axis plots the classification error rate for the features derived by each method. We can see that our NODL method (blue) yields lower errors than the baseline ODL (magenta) for relatively small subsets of features, although the difference is negligible for the full feature set. Overall, this suggests that our NODL approach achieves better reconstruction performance of the input data, without extra overfitting in classification setting, since it generalizes at least as good as, and often better than the baseline ODL method.

**Non-sparse dictionary elements**
When exploring a wide range of sparsity settings (see Appendix), we observed quite different results for non-sparse dictionaries as opposed to those presented above. Fig. 8(b) (in Appendix, due to space constraints) summarizes the results for a particular setting of fully dense dictionaries (no zero entries), but sparse codes (50 non-zeros out of up to 600 dictionary elements; however, the codes are still dense when dictionary size is below 50). In this setting, unlike the previous one, we do not observe any significant improvement in accuracy due to neurogenetic approach, neither in reconstruction nor in classification accuracy; both methods perform practically the same. (Also, note a somewhat surprising phenomenon: after a certain point, i.e. about 50 elements, the reconstruction accuracy of both methods actually declines rather than improves with increasing dictionary size.)

[7] In our experiments, we also track which dictionary elements are deleted by our method; generally, both old and newly added elements get deleted, depending on specific settings.

It is interesting to note, however, that the overall classification errors, for both methods, are much higher in this setting (from 0.4 to 0.52) than in the sparse-dictionary setting (from 0.22 to 0.36). Even using non-sparse codes in the non-sparse dictionary setting still yields inferior results when compared to sparse dictionaries (see the results in the Appendix).

*In summary, on real-life image datasets we considered herein, our NODL approach is often superior (and never inferior) to the standard ODL method; also, there is a consistent evidence that our approach is most beneficial in sparse dictionary settings.*

### 4.2 Sparse Orthogonal Inputs: NLP and Synthetic Data

So far, we explored some conditions on methods properties (e.g., sparse versus dense dictionaries, as well as code sparsity/density) which can be beneficial for the neurogenetic approach. Our further question is: what kind of specific data properties would best justify neurogenetic versus traditional, fixed-size dictionary learning? As it turns out, the fixed-size ODL approach has difficulties adapting to a new domain in nonstationary settings, when the data in both domains are sparse and, across the domains, the supports (i.e., the sets of non-zero coordinates) are almost non-overlapping (i.e., datasets are nearly orthogonal). This type of data properties is related to a natural language processing problem considered below. Furthermore, pushing this type of structure to the extreme, we used simulations to better understand the behavior of our method. Herein, we focused, again, on sparse dictionary elements, as a well-suited basis for representing sparse data. Moreover, our empirical results confirm that using dense dictionary elements does not yield good reconstruction of sparse data, as expected.

**Sparse Natural Language Processing Problem**
We consider a very sparse word co-occurrence matrix (on average, about 14 non-zeros in a column of size 12,883) using the text from two different domains, biology and mathematics, with the total vocabulary size of approximately 12,883 words. The full matrix was split in two for illustration purposes and shown in Fig. 4(c) and 4(d), where math terms correspond to the first block of columns and the biology terms correspond to the second one (though it might be somewhat hard to see in the picture, the average number of nozeros per row/column is indeed about 14).

We use the sparse columns (or rows) in the matrix, indexed by the vocabulary words, as our input data to learn the dictionary of sparse elements (25 non-zeros) with sparse codes (38 non-zeros). The corresponding word codes in the learned dictionary can be later used as word embeddings, or word vectors, in various NLP tasks such as information extraction, semantic parsing, and others Yogatama et al. (2015); Faruqui et al. (2015); Sun et al. (2016). (Note that many of the non-domain specific words were removed from the vocabulary to obtain the final size of 12,883.) Herein, we evaluate our NODL method (i.e. NODL (sparse) in the plots) versus baseline ODL dictionary learning approach (i.e. ODL (sparse)) in the settings where the biology domain is processed first and then one have to switch to the the mathematics domain. We use 2750 samples from each of the domains for training and the same number for testing. The evaluation results are shown in Fig. 4. For the first domain (biology), both methods perform very similarly (i.e., remember the old data equally well), while for the second, more recent domain, our NODL algorithm is clearly outperforming its competitor. Moreover, as we mention above, non-sparse (dense) dictionaries are not suited for the modeling of highly sparse data such as our NLP data. In the Fig. 4, both random dense dictionaries (random-D) and the dense dictionaries learned with ODL (i.e. ODL (dense)) do poorly in the biology and mathematics domains.

However, the reconstruction accuracy as measured by Pearson correlation was not too high, overall, i.e. the problem turned out to be more challenging than encoding image data. It gave us an intuition about the structure of sparse data that may be contributing to the improvements due to neurogenesis. Note that the word co-occurrence matrix from different domains such as biology and mathematics tends to have approximately block-diagonal structure, where words from the same domain are occurring together more frequently than they co-occur with the words from the different domain. Pushing this type of structure to extreme, we studied next the simulated sparse dataset where the samples from the two different domains are not only sparse, but have completely non-overlapping supports, i.e. the data matrix is block-diagonal (see Fig. 7(a) in Appendix).

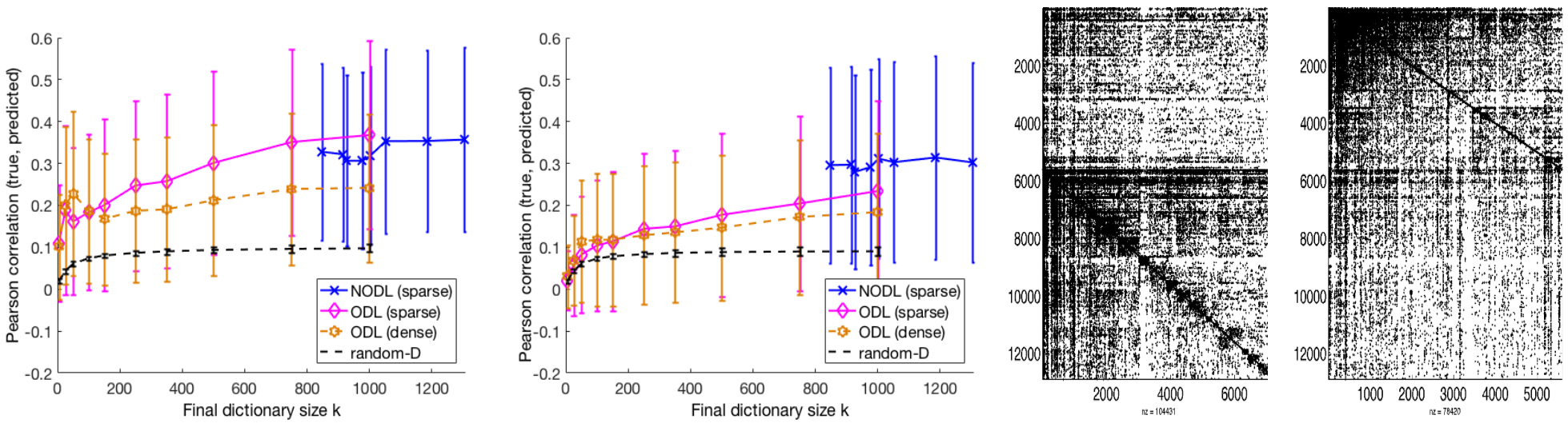

(a) 1st domain (Biology) (b) 2nd Domain (Mathematics) (c) Biology (d) Math

Figure 4: Reconstruction accuracy for the sparse NLP data.

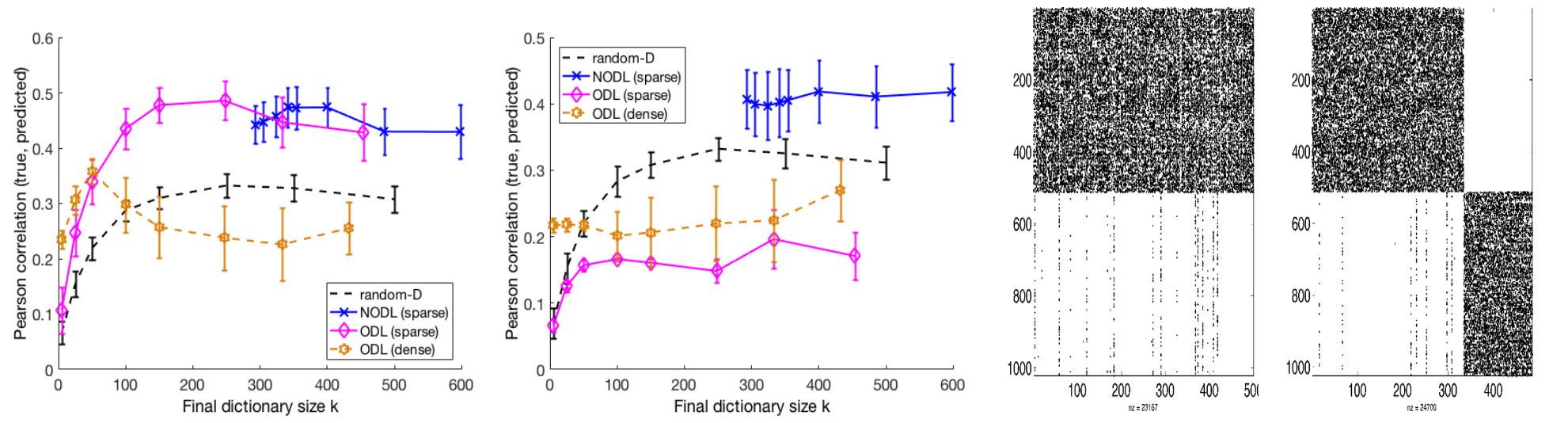

(a) Pearson- First Domain (b) Pearson- Second Domain (c) $\boldsymbol{D}$- ODL (d) $\boldsymbol{D}$- NODL (ours)

Figure 5: Reconstruction accuracy for the sparse synthetic data.

**Synthetic Sparse Data**

We generated a synthetic sparse dataset with 1024 dimension, and only 50 nonzeros in each sample. Moreover, we ensured that the data in both domains had non-overlapping supports (i.e., non-intersecting sets of non-zero coordinates), by always selecting nonzeros in the first domain from the first 512 dimensions, while only using the last 512 dimensions for the second domain Fig. 7(a) in Appendix). For the evaluation on the synthetic data, we use the total of 200 samples for the training and testing purposes each (100 samples for each of the two domains), and smaller batches for online training, containing 20 samples each (instead of 200 samples used earlier for images and language data).

Since the data is sparse, we accordingly adjust the sparsity of dictionary elements (50 nonzeros in an element; for the code sparsity, we will present the results with 50 nonzeros as well). In Fig. 5, we see reconstruction accuracy, for the first and second domain data. For the first domain, the baseline ODL method (i.e. ODL (sparse) in the plots) and our NODL (i.e. NODL (sparse)) perform equally well. On the other hand, for the second domain, the ODL algorithm's performance degrades significantly compared to the first domain. This is because the data from the second domain have non-overlapping support w.r.t. the data from the first domain. Our method is able to perform very well on the second domain (almost as good as the first domain). It is further interesting to analyze the case of random non-sparse dictionary (random-D) which even performs better than the baseline ODL method, for the second domain. This is because random dictionary elements remain non-sparse in all the dimensions thereby doing an average job in both of the domains. Along the same lines, ODL (dense) performs better than the ODL (sparse) in the second domain. Though, the performance of non-sparse dictionaries should degrade significantly with an increase in the sparsity of data, as we see above for the NLP data. Clearly, our NODL (sparse) gives consistently better reconstruction accuracy, compared to the other methods, across the two domains.

In Fig. 5(c) and Fig. 5(d), we see the sparsity structure of the dictionary elements learned using the baseline ODL method and our NODL method respectively. From these plots, we get better insights on why the baseline method does not work. It keeps same sparsity structure as it used for the data from the first domain. Our NODL adapts to the second domain data because of its ability to add new dictionary elements, that are randomly initialized with non-zero support in all the dimensions.

Next, in Sec. 5, we discuss our intuitions on why NODL performs better than the ODL algorithm under certain conditions.

## 5 WHEN NEUROGENESIS CAN HELP, AND WHY

In the Sec. 4, we observed that our NODL method outperforms the ODL algorithm in two general settings, both involving sparse dictionary elements: (i) non-sparse data such as real-life images, and (ii) sparse data with (almost) non-overlapping supports. In this section, we attempt to analyze what contributes to the success of our approach in these settings, starting with the last one.

**Sparse data with non-overlapping supports, sparse dictionary**
As discussed above, in this scenario, the data from both the first and the second domain are sparse, and their supports (non-zero dimensions) are non-overlapping, as shown in the Fig. 7(a). Note that, when training a dictionary using the fixed-size, sparse-dictionary ODL method, we observe only a minor adaptation to the second domain after training on the first domain, as shown in Fig. 5(c).

Our empirical observations are supported by the theoretical result summarized in Lemma 1 below. Namely, we prove that when using the ODL algorithm in the above scenario, the dictionary trained on the first domain can not adapt to the second domain. (The minor adaptation, i.e., a few nonzeros, observed in our results in Fig. 5(c) occurs only due to implementation details involving normalization of sparse dictionary elements when computing codes in the dictionary – the normalization introduces non-zeros of small magnitude in all dimensions (see Appendix for the experiment results with no normalization of the elements, conforming to the Lemma 1)).

**Lemma 1.** *Let $\boldsymbol{x}_1, \boldsymbol{x}_2, \cdots, \boldsymbol{x}_{t-1} \in \mathbb{R}^m$ be a set of samples from the first domain, with non-zeros (support) in the set of dimensions $P \subset M = \{1, \cdots, m\}$, and let $\boldsymbol{x}_t, \boldsymbol{x}_{t+1}, \cdots, \boldsymbol{x}_n \in \mathbb{R}^m$ be a set of samples from the second domain, with non-zeros (support) in dimensions $Q \subset M$, such that $P \cap Q = \emptyset$, $|P| = |Q| = l$. Let us denote as $\boldsymbol{d}_1, \boldsymbol{d}_2, \cdots, \boldsymbol{d}_k \in \mathbb{R}^m$ dictionary elements learned by ODL algorithm, with the sparsity constraint of at most $l$ nonzeros in each element [8], on the data from the first domain, $\boldsymbol{x}_1, \cdots, \boldsymbol{x}_{t-1}$. Then (1) those elements have non-zero support in $P$ only, and (2) after learning from the second domain data, the support (nonzero dimensions) of the corresponding updated dictionary elements will remain in $P$.*

*Proof Sketch.* Let us consider processing the data from the first domain. At the first iteration, a sample $\boldsymbol{x}_1$ is received, its code $\boldsymbol{\alpha}_1$ is computed, and the matrices $\boldsymbol{A}$ and $\boldsymbol{B}$ are updated, as shown in Alg. 1 (non-highlighted part); next, the dictionary update step is performed, which optimizes

$$\boldsymbol{D}^{(1)} = \arg\min_{\boldsymbol{D} \in \mathcal{C}} \frac{1}{2} Tr(\boldsymbol{D}^T \boldsymbol{D} \boldsymbol{A}) - Tr(\boldsymbol{D}^T \boldsymbol{B}) + \sum_j \lambda_j ||\boldsymbol{d}_j||_1. \quad (6)$$

Since the support of $\boldsymbol{x}_1$ is limited to $P$, we can show that optimal dictionary $\boldsymbol{D}^*$ must also have all columns/elements with support in $P$. Indeed, assuming the contrary, let $\mathbf{d_j}(i) \neq 0$ for some dictionary element/column $j$, where $i \notin P$. But then it is easy to see that setting $\mathbf{d_j}(i)$ to zero reduces the sum-squared error and the $l_1$-norm in (6), yielding another dictionary that achieves a lower overall objective; this contradicts our assumption that $\mathbf{D}^*$ was optimal. Thus, the dictionary update step must produce a dictionary where all columns have their support in $P$. By induction, this statement will also be true for the dictionary obtained after processing all samples from the first domain. Next, the samples from the second domain start arriving; note that those samples belong to a different subspace, spanning the dimensions within the support set $Q$, which is not intersecting with $P$. Thus, using the current dictionary, the encoding $\boldsymbol{\alpha}_t$ of first sample $\boldsymbol{x}_t$ from the second domain (i.e. the solution of the LASSO problem in step 4 of the Alg. 1 ) will be a zero vector. Therefore, the matrices $\boldsymbol{A}$ and $\boldsymbol{B}$ remains unchanged during the update in step 11, and thus the support of each $\boldsymbol{b}_j$, and, consequently, $\boldsymbol{u}_j$ and the updated dictionary elements $\boldsymbol{d}_j$ will remain in $P$. By induction, every dictionary update in response to a new sample from the second domain will preserve the support of the dictionary elements, and thus the final dictionary elements will also have their support only in $P$.

□

**Non-sparse data, sparse dictionary**
We will now discuss an intuitive explanation behind the success of neurogenetic approach in this scenario, leaving a formal theoretical analysis as a direction for future work. When learning sparse

[8] $l$ corresponds to $\beta_d$ in Alg. 1

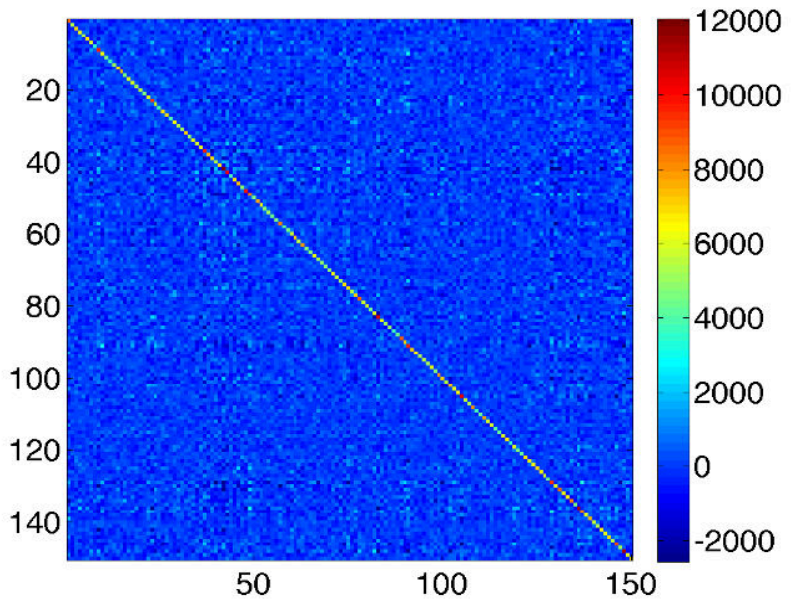

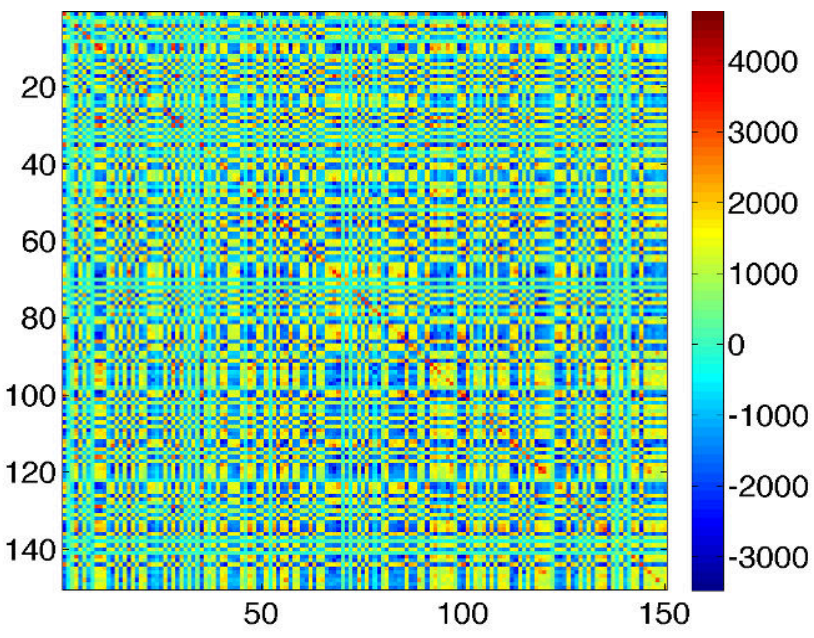

(a) $\boldsymbol{A}$ with ODL method (with dense elements) (b) $\boldsymbol{A}$ with ODL method (with sparse elements)

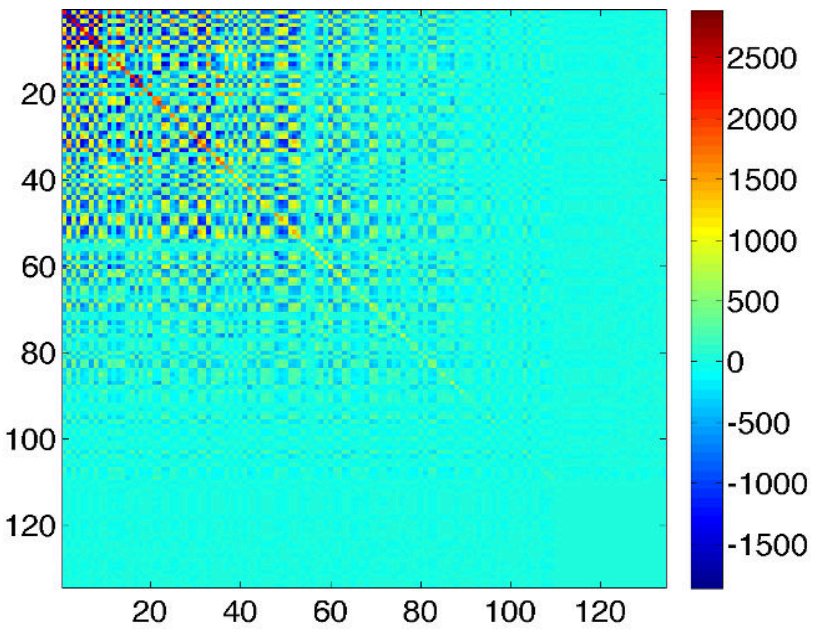

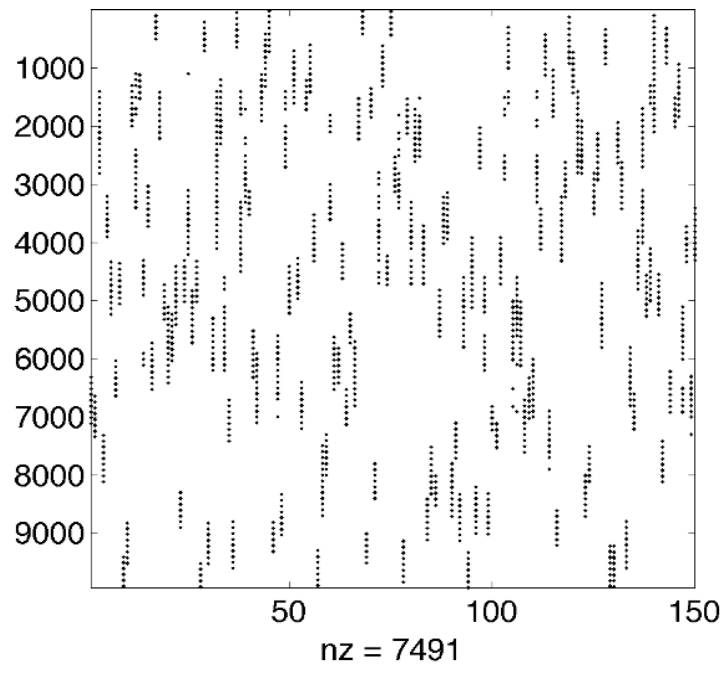

(c) $\boldsymbol{A}$ with our method (with sparse elements) (d) $\boldsymbol{D}$ with ODL method (with sparse elements)

Figure 6: Visualization of the sparse dictionary and the matrix $\mathbf{A}$ learned on the first imaging domain (Oxford images), using the baseline ODL method and our method.

dictionaries on non-sparse data such as natural images, we observed that many dictionary elements have non-overlapping supports with respect to each other; see, for example, Fig. 6(d), where each column corresponds to a 10000-dimensional dictionary element with nonzero dimensions shown in black color. Apparently, the non-zeros dimensions of an element tend to cluster spatially, i.e. to form a patch in an image. The non-overlapping support of dictionary elements results into a specific structure of the matrix $\boldsymbol{A}$. As shown in Fig. 6(b), for ODL approach, the resulting matrix $\boldsymbol{A}$ includes many off-diagonal nonzero elements of large absolute values (along with high values on the diagonal). Note that, by definition, $\boldsymbol{A}$ is an empirical covariance of the code vectors, and it is easy to see that a nonzero value of $a_{jk}$ implies that the $j$-th and the $k$-th dictionary elements were used jointly to explain the same data sample(s). Thus, the dense matrix structure with many non-zero off-diagonal elements, shown in Fig. 6(b), implies that, when the dictionary elements are sparse, they will be often used jointly to reconstruct the data. On the other hand, in the case of non-sparse dictionary elements, the matrix $\boldsymbol{A}$ has an almost diagonally-dominant structure, i.e. only a few dictionary elements are used effectively in the reconstruction of each data sample even with non-sparse codes (see Appendix for details).

Note that in the dictionary update expression $\boldsymbol{u}_j \leftarrow \frac{\boldsymbol{b}_j - \sum_{k \neq j} \boldsymbol{d}_k a_{jk}}{a_{jj}}$ in (3), when the values $a_{jk}/a_{jj}$ are large for multiple $k$, the $j_{th}$ dictionary element becomes tightly coupled with other dictionary elements, which reduces its adaptability to new, non-stationary data. In our algorithm, the values $a_{jk}/a_{jj}$ remain high if both elements $j$ and $k$ have similar “age”; however, those values are much lower if one of the elements is introduced by neurogenesis much more recently than the other one. In 6(c), the upper left block on the diagonal, representing the oldest elements (added during the initialization), is not diagonally-dominant (see the sub-matrices of $\boldsymbol{A}$ with NODL in Fig. 14 in the Appendix). The lower right block, corresponding to the most recently added new elements, may also have a similar structure (though not visible due to relatively low magnitudes of the new elements; see the Appendix). Overall, our interpretation is that the old elements are tied to each other whereas the new elements may also be tied to each other but less strongly, and not tied to the old elements, yielding a block-diagonal structure of $\mathbf{A}$ in case of neurogenetic approach, where blocks correspond to dictionary elements adapted to particular domains. In other words, neurogenesis allows for an adaptation to a new domain without forgetting the old one.

## 6 Conclusions

In this work, we proposed a novel algorithm, Neurogenetic Online Dictionary Learning (NODL), for the problem of learning representations in non-stationary environments. Our algorithm builds a dictionary of elements by learning from an online stream of data while also adapting the dictionary structure (the number of elements/hidden units and their connectivity) via continuous birth (addition) and death (deletion) of dictionary elements, inspired by the adult neurogenesis process in hippocampus, which is known to be associated with better adaptation of an adult brain to changing environments. Moreover, introducing sparsity in dictionary elements allows for adaptation of the hidden unit connectivity and further performance improvements.

Our extensive empirical evaluation on both real world and synthetic data demonstrated that the interplay between the birth and death of dictionary elements allows for a more adaptive dictionary learning, better suited for non-stationary environments than both of its counterparts, such as the fixed-size online method of Mairal et al. (2009) (no addition and no deletion), and the online version of the group-sparse coding method by Bengio et al. (2009) (deletion only). Furthermore we evaluated, both empirically and theoretically, several specific conditions on both method's and data properties (involving the sparsity of elements, codes, and data) where our method has significant advantage over the standard, fixed-size online dictionary learning. Overall, we can conclude that neurogenetic dictionary learning typically performs as good as, and often much better than its competitors. In our future work, we plan to explore the non-linear extension of the dictionary model, as well as a stacked auto-encoder consisting of multiple layers.

## A IMPLEMENTATION DETAILS

In our implementation of a sparsity constraint with a given number of non-zeros, we perform a binary search for the value of the corresponding regularization parameter, $\lambda$, as shown in Alg. 2. This approach costs much lesser than other techniques such as LARS while the quality of solutions are very similar.

**Algorithm 2** Binary search of $\lambda$ with the proximal method based sparsity

**Require:** $\boldsymbol{u}$ (vector to be sparsified) , $\beta$ (numbers of non-zeros),
$\epsilon_\beta$ (acceptable error in $\beta$), $\epsilon_\lambda$ (acceptable error in $\lambda$)

$\boldsymbol{u}^+ = abs(\boldsymbol{u})$
$\lambda_{min} = 0$ (if no sparsity)
$\lambda_{max} = max(\boldsymbol{u}^+)$
**while** true **do**
  $\lambda_{mean} = \frac{\lambda_{min}+\lambda_{max}}{2}$
  $\beta^* = nnz((\boldsymbol{u}^+ - \lambda_{mean})_+)$ (non zeros with proximal operator)
  **if** $\frac{abs(\lambda_{max}-\lambda_{min})}{\lambda_{max}} < \epsilon_\lambda$ or $abs(\beta^* - \beta) \leq \epsilon_\beta$ **then**
    $\lambda = \lambda_{mean}$
    **return** $\lambda$
  **else if** $\beta^* > \beta$ **then**
    $\lambda_{min} = \lambda_{mean}$
  **else if** $\beta^* < \beta$ **then**
    $\lambda_{max} = \lambda_{mean}$
  **else**
    error: this condition is not possible.
  **end if**
**end while**

# B Experimental Results

## B.1 Additional plots for the experiment results discussed in Sec. 4

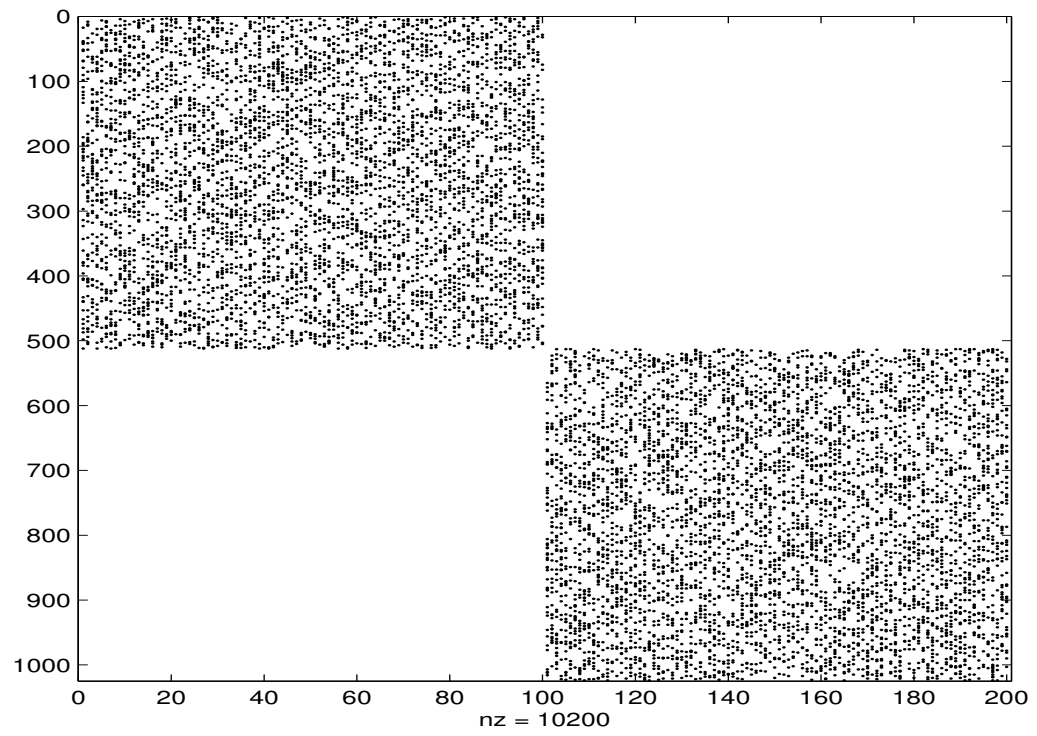

(a) Synthetic data

Figure 7: The data sets for the evaluation of the online dictionary learning algorithms.

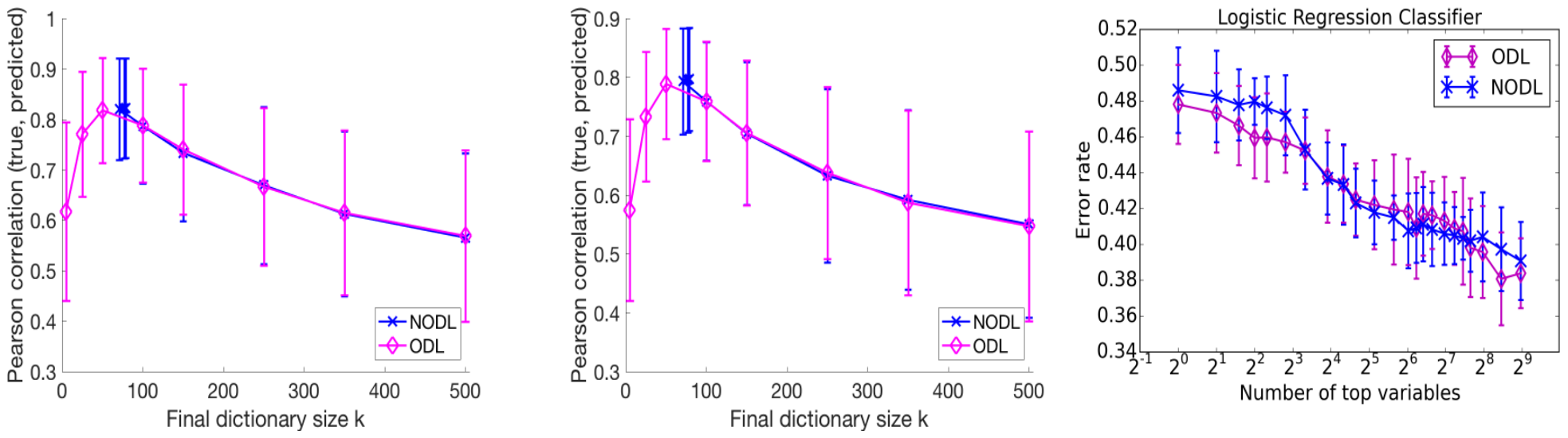

(a) 1st domain (Oxford)
(b) 2nd domain (Flowers)
(c) Classification Error

Figure 8: Reconstruction accuracy for 100x100 size images with non-sparse dictionary but sparse code (50 non-zeros) settings.

Fig. 9 is an extension of the original Fig. 2 in the main paper, for the same experiments, on the images compressed to size 32x32, in the learning settings of sparse dictionary elements, and relatively less sparse codes. This figure is included to show that: (i) our analysis for the results presented in Fig. 2 extends to the other metric, mean square error (MSE); (ii) the results for the second domain data of flowers and animals are also highly similar to each other. Along the same lines, Fig. 10, Fig. 11 extend Fig. 3, Fig. 8 respectively. In these extended figures, we see the similar generalizations across the evaluation metrics, the data sets and the classifiers.

## B.2 Trade off between the sparsity of dictionary elements and codes

In this section, our results from the experiments on 32x32 size compression of images are presented where we vary the sparsity of codes as well as the sparsity of dictionary elements for a further analysis of the trade off between the two. From the left to right in Fig. 12, we keep dictionary elements sparse but slowly decrease their sparsity while increasing the sparsity of codes. Here, the number of nonzeros in a dictionary element (dnnz) and a code (cnnz) are decided such that it produces the overall number of non-zeros approx. to the size of an image (i.e. 32x32) if there were no overlaps of non-sparse patches between the elements. We observe the following from the figure: (i) the overall reconstruction gets worse when we trade off the sparsity of dictionary elements for the sparsity of codes. (ii) the performance of our NODL method is better than the baseline ODL method, especially when there is higher sparsity in dictionary elements.

### B.3 Non-sparse dictionary elements and non-sparse codes

We also performed experiments for the settings where dictionary elements and codes are both non-sparse. See Fig. 13. For this scenario, while we get very high reconstruction accuracy, the overall classification error remains much higher (ranging between 0.48 to 0.32) compared to the sparse dictionary elements setting in Fig. 10 (0.36 to 0.22), though lower than the settings of non-sparse dictionary with sparse codes in Fig. 11 (0.52 to 0.40).

### B.4 Addition plots for the analysis of sparse dictionary elements

Fig. 14 extends Fig. 6. For the case of non-sparse dictionary elements, the structure of matrix $\boldsymbol{A}$ with the ODL algorithm after processing the first domain image data (oxford images) is shown in Fig. 14(a) for non-sparse codes settings (similar structure for sparse code settings). In both cases of non-sparse elements, sparse codes as well as non-sparse codes, the matrix is diagonally dominant, in contrast to the scenario of sparse dictionary elements in Fig. 6(b). For our algorithm NODL in the settings of sparse dictionary elements, we show the matrix $\boldsymbol{A}$ in Fig. 6(c) and its sub-matrices in Fig. 14(b), 14(c) and 14(d). Fig. 14(b) demonstrates that the old dictionary elements are tied to each other (i.e. high values of $\frac{a_{jk}}{a_{jj}} \forall k \neq j$). Similar argument applies to the recently added new dictionary elements, as in Fig. 14(d), though the overall magnitude range is smaller compared to the old elements in Fig. 14(b). Also, we see that the new elements are not as strongly tied to each other as the old elements, but more than the case of non-sparse dictionary elements. In Fig. 14(c), we can see more clearly that the new elements are not tied to the old elements. Overall, from the above plots, our analysis is that the new elements are more adaptive to the new non-stationary environments as those new elements are not tied to the old elements, and only weakly tied to each other.

### B.5 Synthetic Sparse data settings

For the case of modeling the synthetic sparse data with sparse dictionary elements, Fig. 15 extends Fig. 5 with the plots on the other metric, mean square error (MSE). In this figure, the ODL algorithm adapts to the second domain data though not as good as our algorithm NODL. Even this adaptation of ODL is due to the normalization of dictionary elements, when computing codes, as we mention in the main draft. If there is no normalization of dictionary elements, the ODL algorithm doesn't adapt to the second domain data at all. For these settings, the results are shown in Fig. 16.

### B.6 Reconstructed Images

In Fig. 17, 18, we show the reconstruction, for some of the randomly picked images from the animals data set, with sparse dictionary elements, and non-sparse elements respectively (500 elements). We suggest to view these reconstructed images in the digital version to appreciate the subtle comparisons. For the case of non-sparse elements in Fig. 18, the reconstructions are equally good for both ODL and our NODL algorithm. On the other hand, for the sparse elements settings, our algorithm NODL gives much better reconstruction than the baseline ODL, as we see visually in Fig. 17. These comparisons of the reconstructed images conform to the evaluation results presented above. It is interesting to see that, with sparse dictionary elements, the background is smoothed out with an animal in the focus in an image, with good reconstruction of the body parts (especially the ones which distinguish between the different species of animals).

Whereas, the non-sparse dictionary does not seem to distinguish between the two, the background and animal in an image; in some of the reconstructed images, it is hard to distinguish an animal from the background. Clearly, the background in an image should lead to noise in features for tasks such as the binary classification considered above (discussed in Sec. 4). This should also explain why we get much better classification accuracy with the use of sparse dictionary elements rather than non-sparse elements. For the scenario of sparse codes with non-sparse dictionary elements, the reconstructed images are even worse; not shown here due to space constraints.

### B.7 Re-initialization of "dead" dictionary elements

In Mairal et al. (2009), it was also noted that, during dictionary updates, some elements may turn into zero-column (i.e., zero $l_2$ norm); those elements were referred to as "dead" elements, since they do not contribute to the data reconstruction task. The fraction of such dead elements elements was typically very small in our experiments with the original ODLmethod (i.e., without the explicit "killing" of the elements via the group sparsity regularization). In Mairal et al. (2009), it is proposed to reinitialize such dead elements, using, for example, the existing batch of data (random values are another option). Here, we will refer to such extension of the baseline ODL method as to ODL*. Specifically, in ODL*, we reinitialize the "dead" elements with random values, and then continue updating them along with the other dictionary elements, on the current batch of data. Fig. 19 extends Fig. 2 with the additional plots including the ODL* extension of the baseline ODLalgorithm, while keeping all experiment settings the same. We can see that there the difference in the performance of ODL and its extension ODL* is negligible, perhaps due to the the fact that the number of the dead elements, without an explicit group-sparsity regularization, is typically very small, as we already mentioned above. We observe that our method outperforms the ODL* version, as well as the original ODL baseline.

### B.8 Evaluating possible effects of varying the order of training datasets

In our original experiments presented in the main paper (Sec. 4), the Oxford buildings images are processed as part of the first domain data set, followed by the mixture of flower and animal images as the second domain data set. One can ask whether a particular sequence of the input datasets had a strong influence on our results; in this section, we will evaluate different permutations of the input data sets. Specifically, we will pick any two out of the three data sets available, and use them as the first and the second domain data, respectively. In Fig. 20, we present test results on the second domain data for the baseline ODL and for our NODL methods, with each subfigure corresponding to one of the six processing orders on data sets used for training. All experimental settings are exactly the same as those used to produce the plots in Fig. 2. Overall, we observe that, for all possible orders of the input datasets, our NODL approach is either superior or comparable to ODL , but never inferior. We see a significant advantage of NODL over ODL when using Oxford or Flowers data sets as the first domain data. However, this advantage is less pronounced when using the Animals data set as the first domain. One possible speculation can be that animal images can be somewhat more complex to reconstruct as compared to the other two types of data, and thus learning their representation first is sufficient for subsequent representation of the other two types of datasets. Investigating this hypothesis, as well as, in general, the effects of the change in the training data complexity, from simpler to more complex or vice versa, where complexity can be measured, for example, as image compressibility, remains an interesting direction for further research.

### B.9 Robustness of our NODL algorithm w.r.t. the tuning parameters

To demonstrate the robustness of our NODL algorithm w.r.t. the tuning parameters, we perform additional experiments by varing each of the tuning parameters, over a wide range of values, while keeping the others same as those used for producing the Fig. 2. In Fig. 21, 22, 23, 24, 25, 26, we vary the tuning parameters $batchsize$, $c_k$, $\lambda_g$, $\beta_c$, $\beta_d$, $\gamma$ respectively, and show the corresponding test results on the flowers dataset of the second domain (see the Alg. 1, in the Sec. 3, for the roles of the tuning parameters in our NODL algorithm). In these plots, we see that our NODL algorithm outperforms the baseline ODL algorithm, consistently across all the parameter settings.

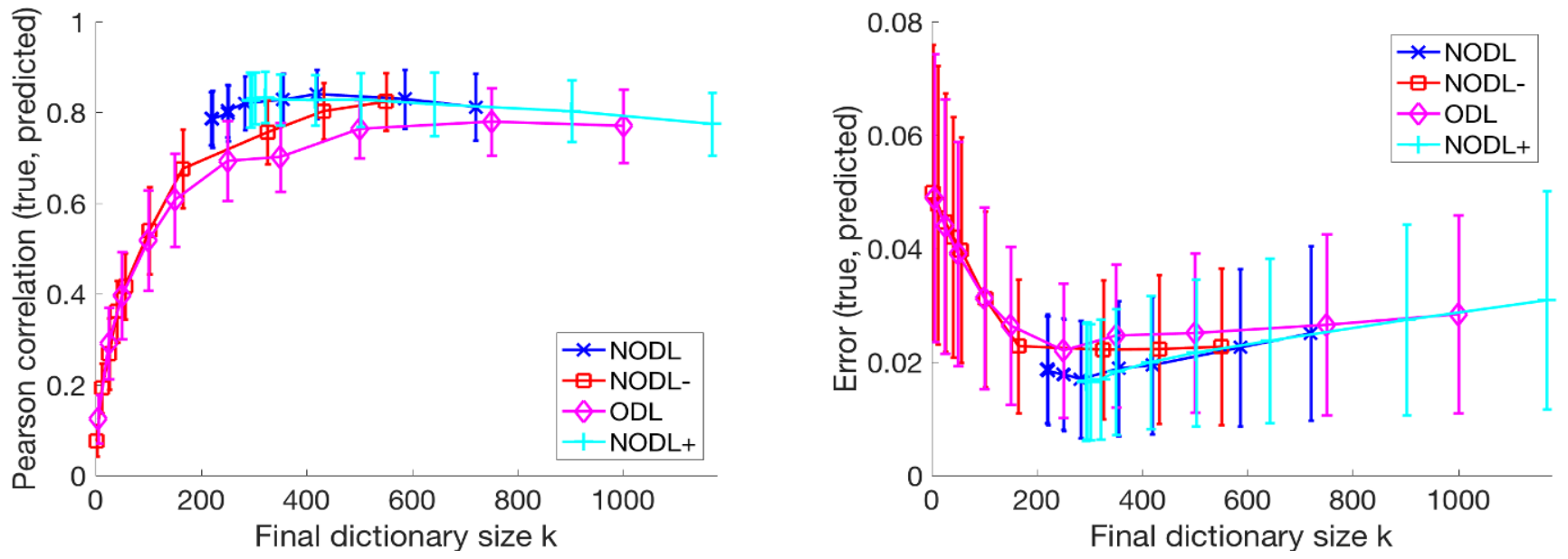

(a) Pearson- Animals (b) MSE- Oxford

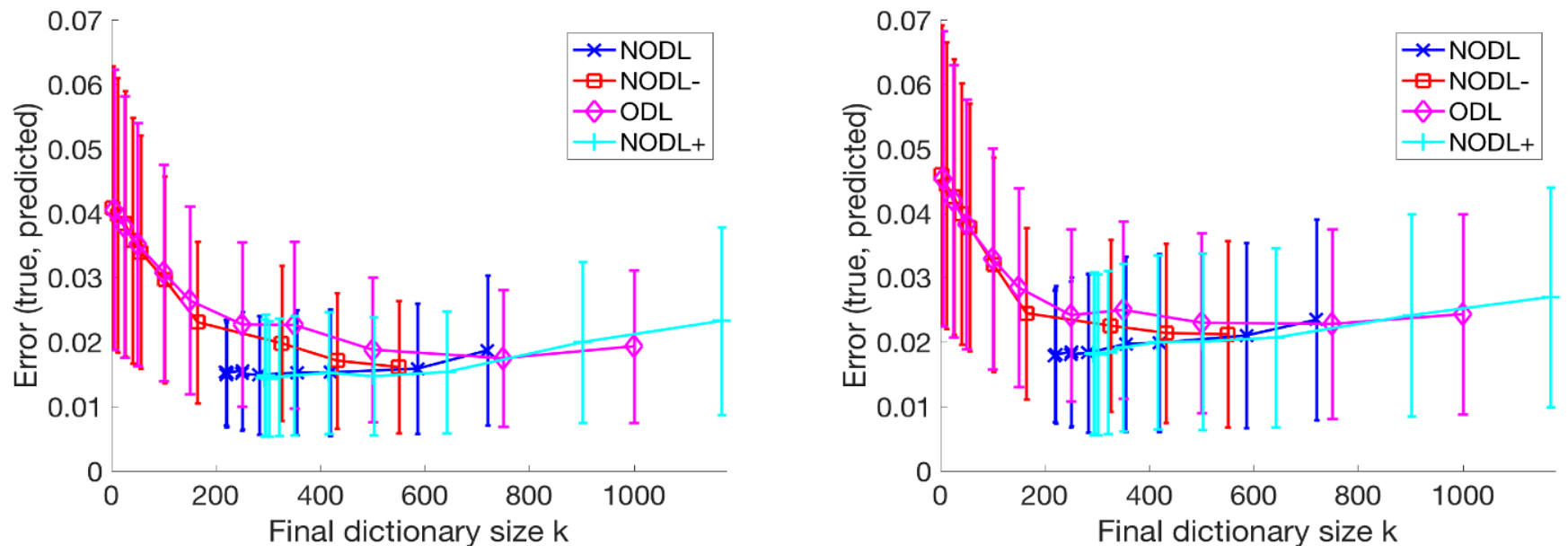

(c) MSE- Flowers (d) MSE- Animals

Figure 9: Reconstruction Error for 32x32 size images with sparse dictionary settings.

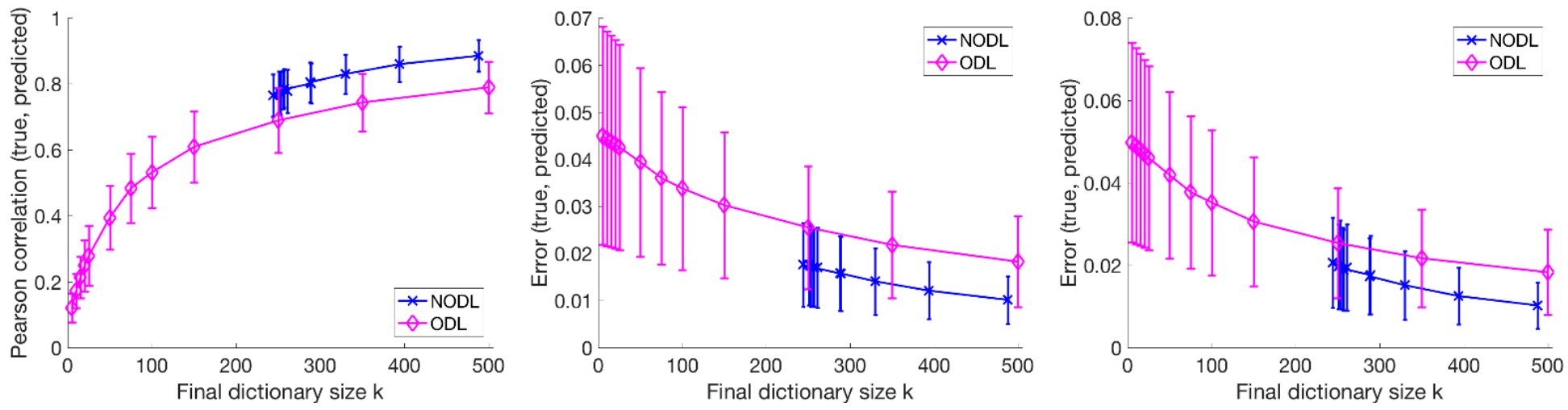

(a) Pearson- Animals (b) MSE- Flowers (c) MSE- Animals

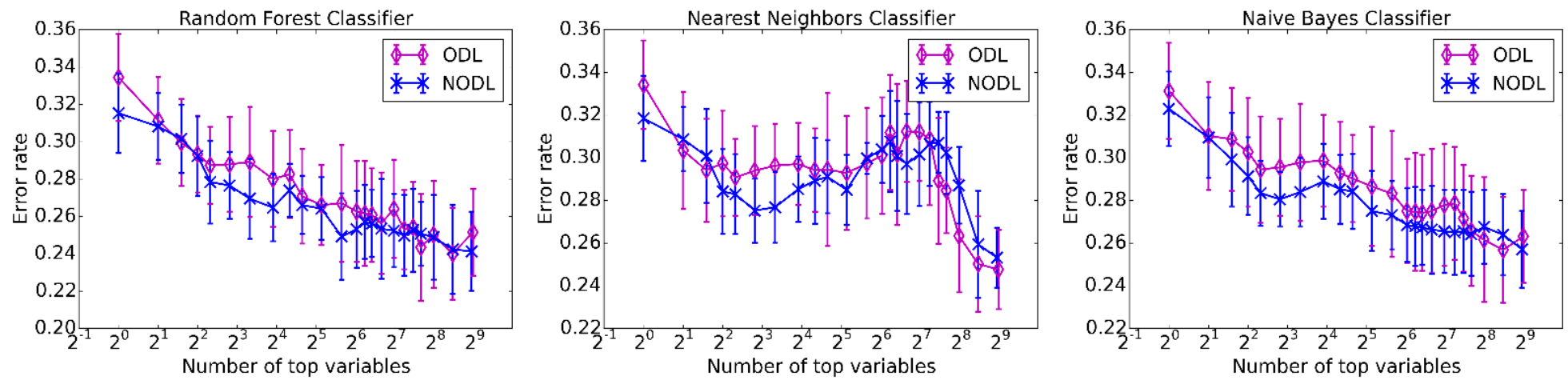

(d) Random Forest (e) Nearest Neighbor (f) Naive Bayes

Figure 10: Reconstruction Error for 100x100 size images with sparse dictionary (50 non-zeros) and non-sparse code settings (2000 non-zeros).

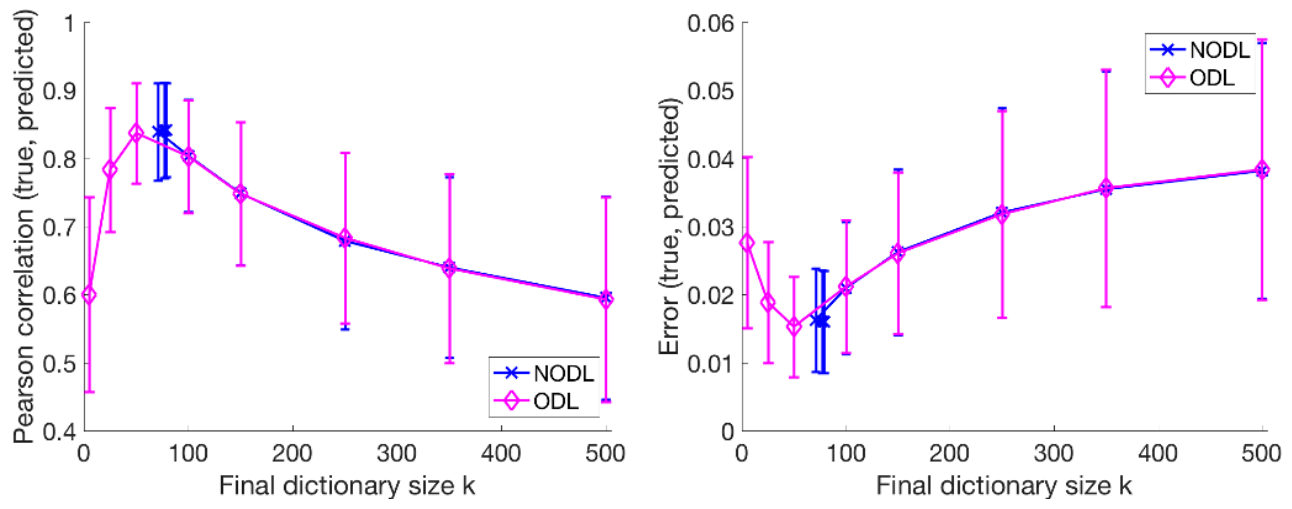

(a) Pearson- Animals (b) MSE- Flowers (c) MSE- Animals

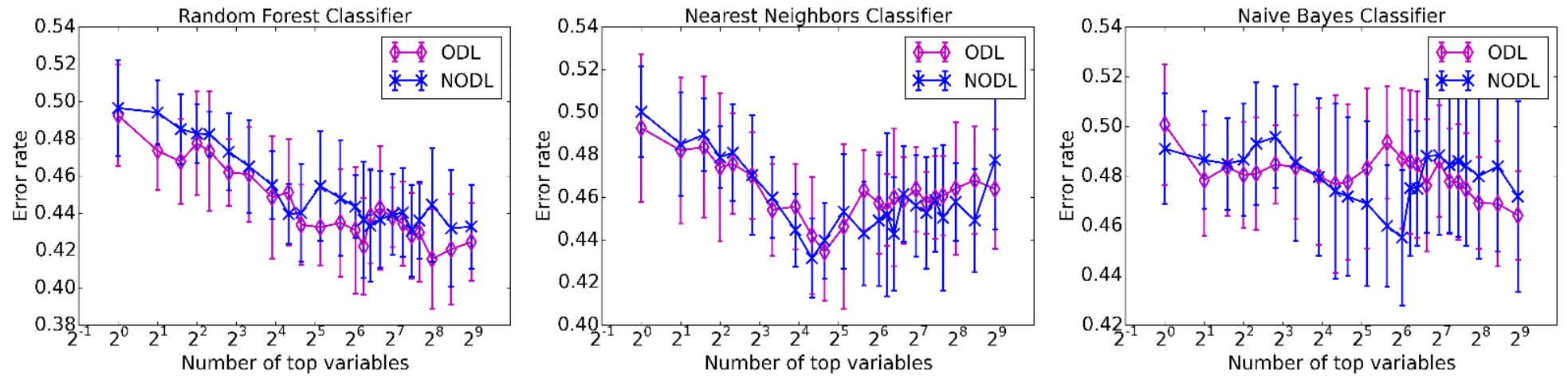

(d) Random Forest (e) Nearest Neighbor (f) Naive Bayes

Figure 11: Reconstruction Error for 100x100 size images with non-sparse dictionary but sparse code (50 non-zeros) settings.

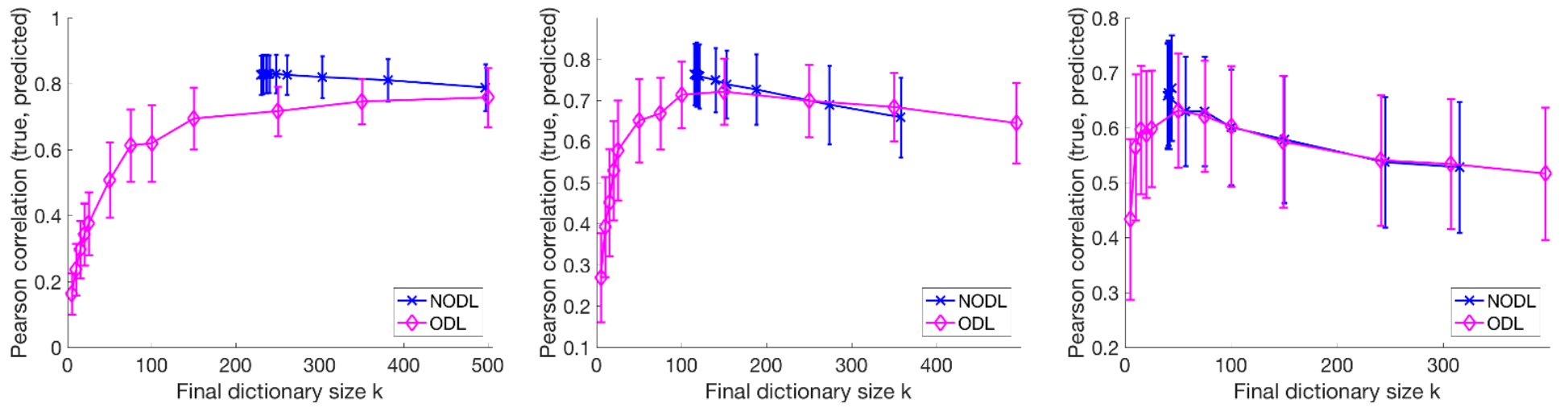

(a) Dnnz:10, Cnnz:100 (Pearson) (b) Dnnz:30, Cnnz:33 (Pearson) (c) Dnnz:100, Cnnz:10 (Pearson)

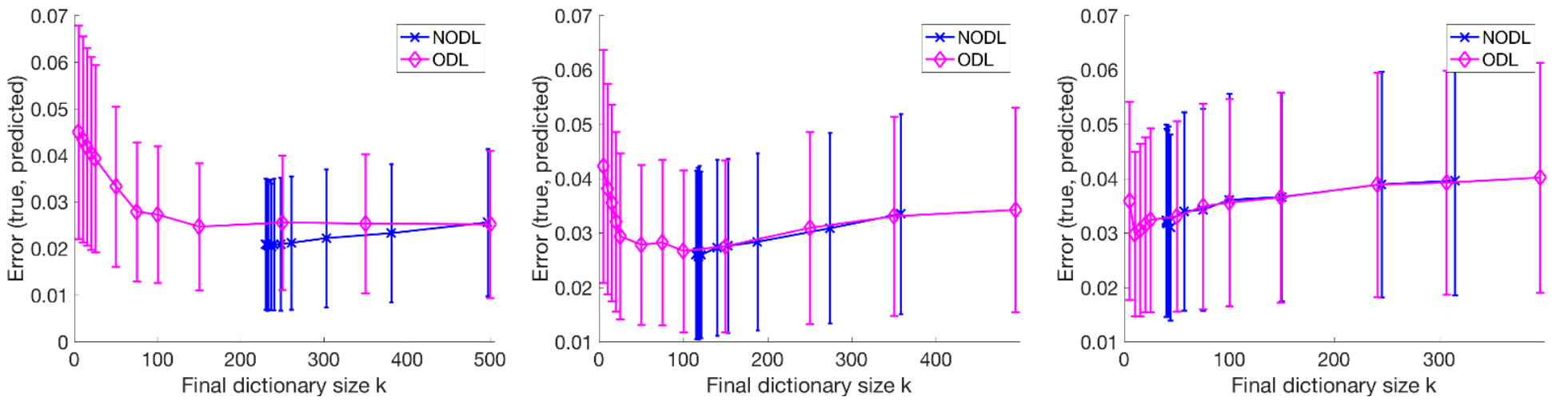

(d) Dnnz:10, Cnnz:100 (MSE) (e) Dnnz:30, Cnnz:33 (MSE) (f) Dnnz:100, Cnnz:10 (MSE)

Figure 12: Reconstruction Error for 32x32 size images, on the animals data, with varying sparsity in dictionary elements and codes.

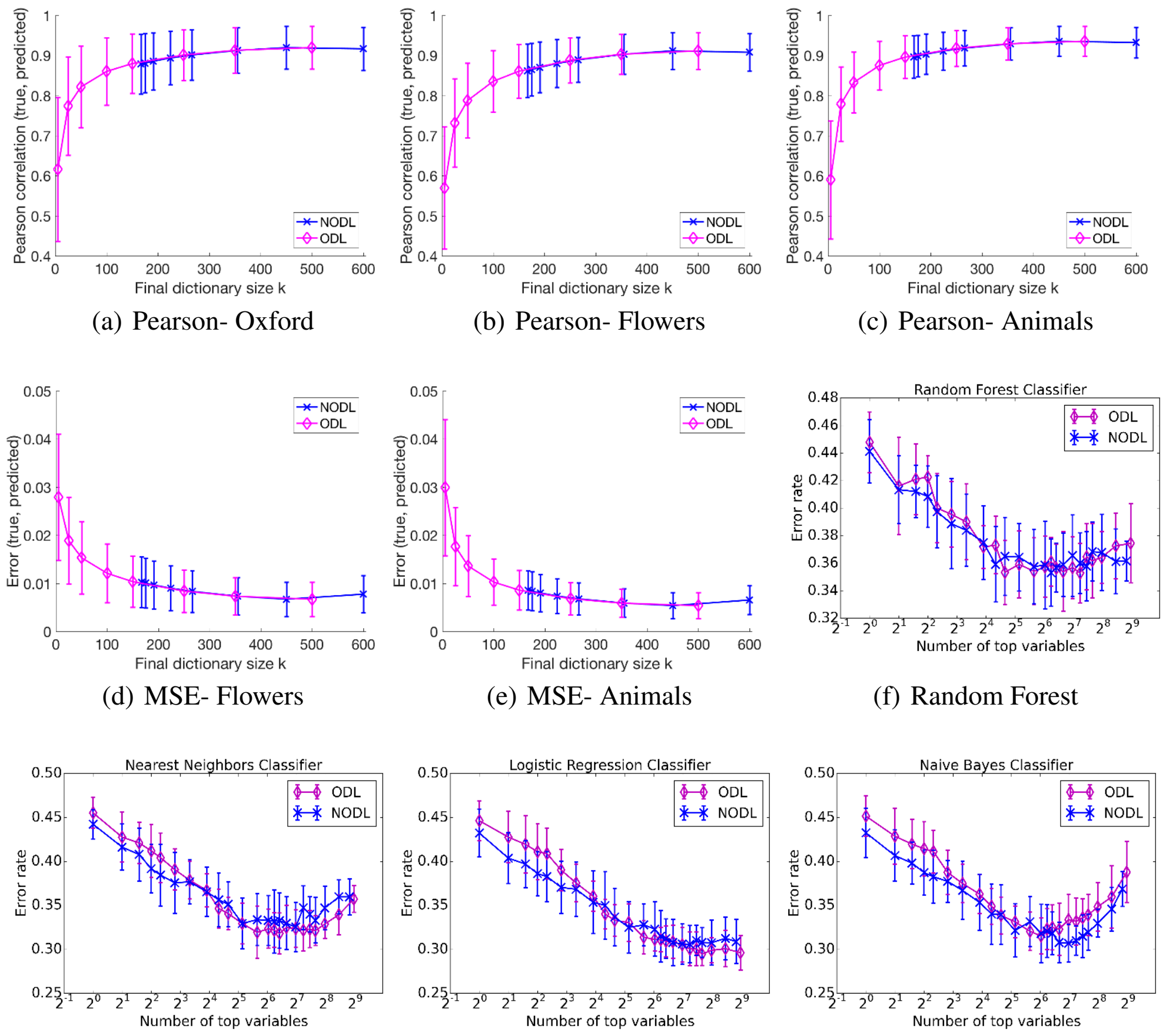

(a) Pearson- Oxford (b) Pearson- Flowers (c) Pearson- Animals

(d) MSE- Flowers (e) MSE- Animals (f) Random Forest

(g) Nearest Neighbor (h) Logistic Regression (i) Naive Bayes

Figure 13: Reconstruction Error for 100x100 size images with non-sparse dictionary and non-sparse codes (500 non-zeros) settings.

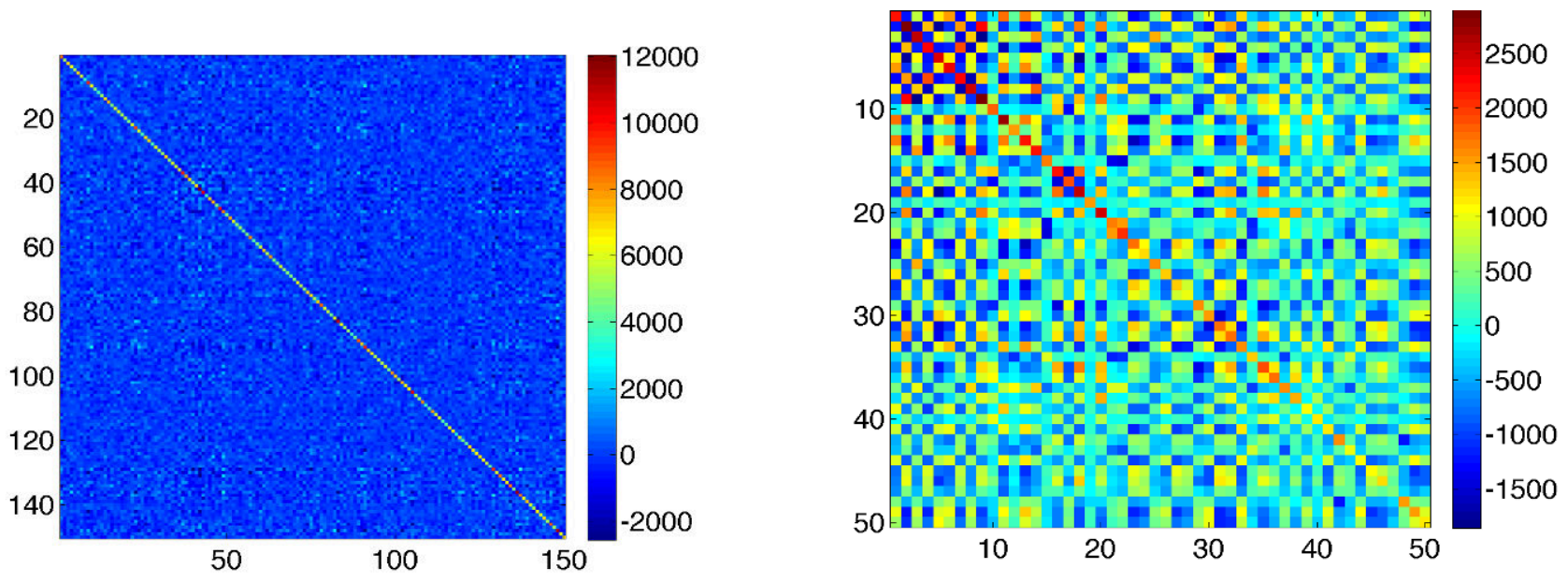

(a) $\boldsymbol{A}$ with ODL method (non-sparse codes, non-sparse dictionary)

(b) $\boldsymbol{A}$ with our method–the old 50 elements (non-sparse codes, sparse dictionary)

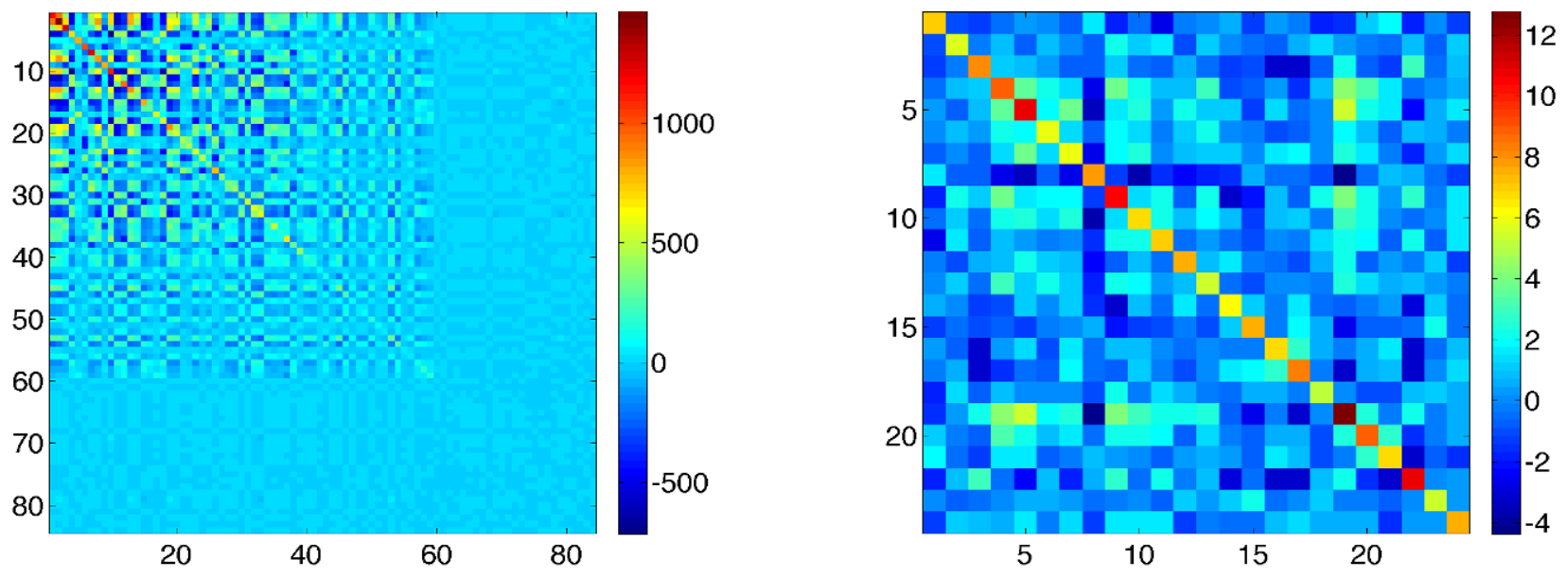

(c) $\boldsymbol{A}$ with our method–all the new elements (non-sparse codes, sparse dictionary)

(d) $\boldsymbol{A}$ with our method–the recently added new elements (non-sparse codes, sparse dictionary)

Figure 14: The structure of a sparse dictionary that is learned from the processing of the first domain image data (Oxford images) using the baseline ODL method.

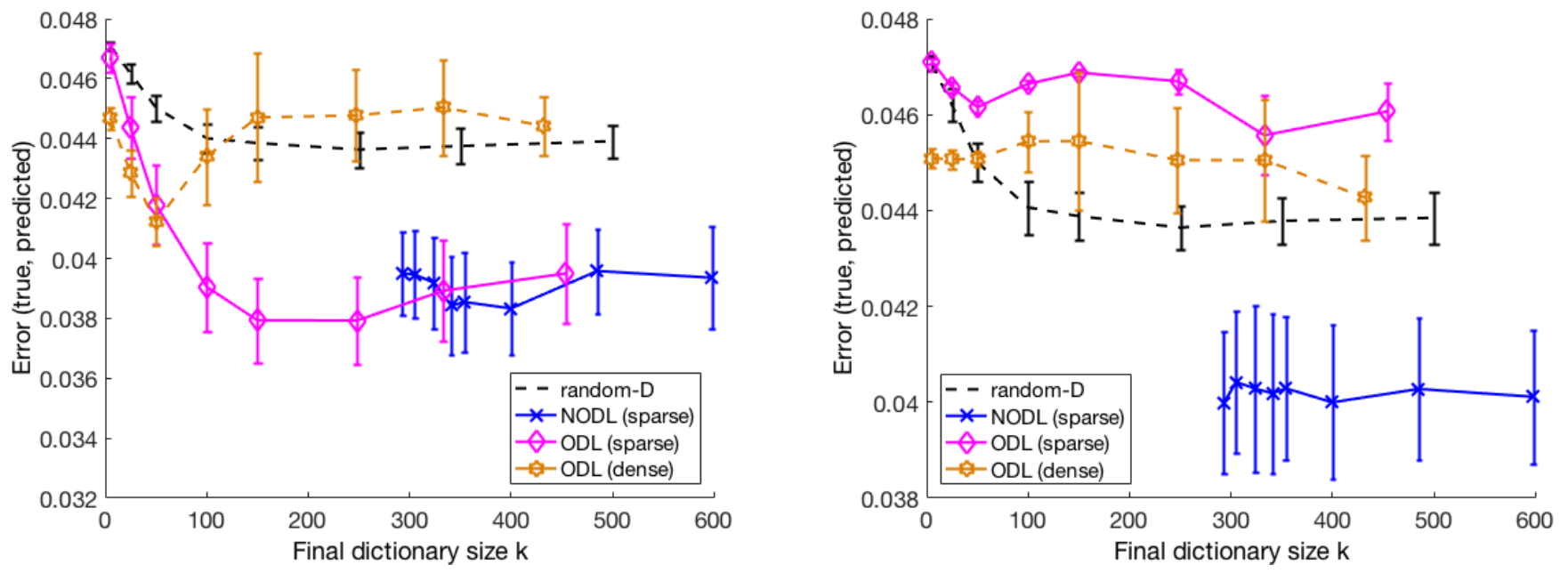

(a) MSE- First Domain

(b) MSE- Second Domain

Figure 15: Reconstruction error for the synthetic data from sub-spaces with non-overlapping supports of non-zeros.

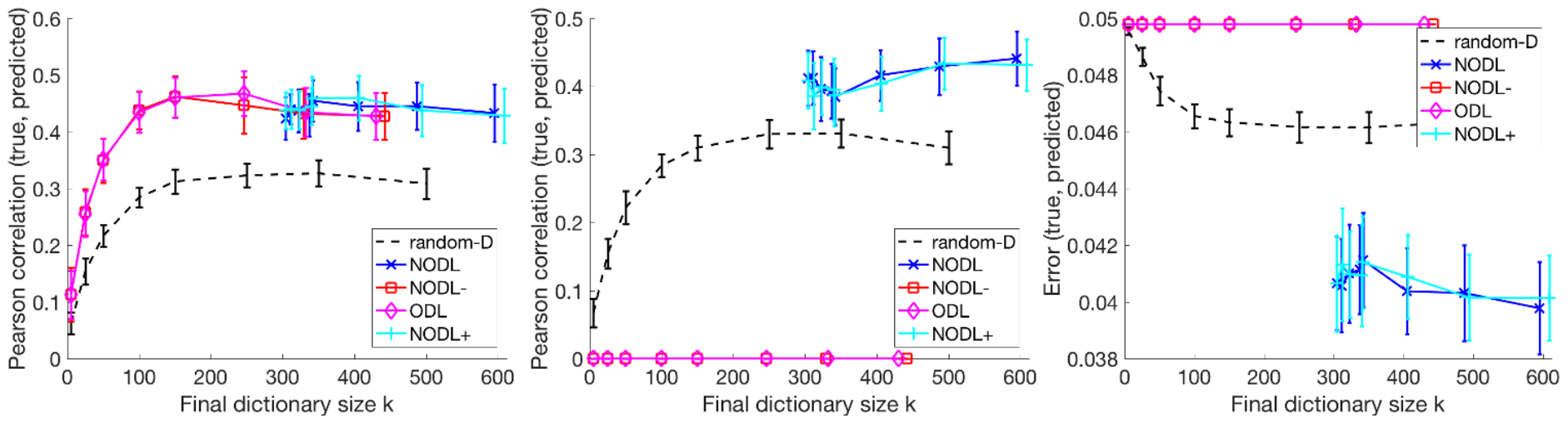

(a) Pearson- First Domain (b) Pearson- Second Domain (c) MSE- Second Domain

Figure 16: Reconstruction error for the synthetic data from sub-spaces with non-overlapping supports of non-zeros (without normalization of dictionary elements when computing codes).

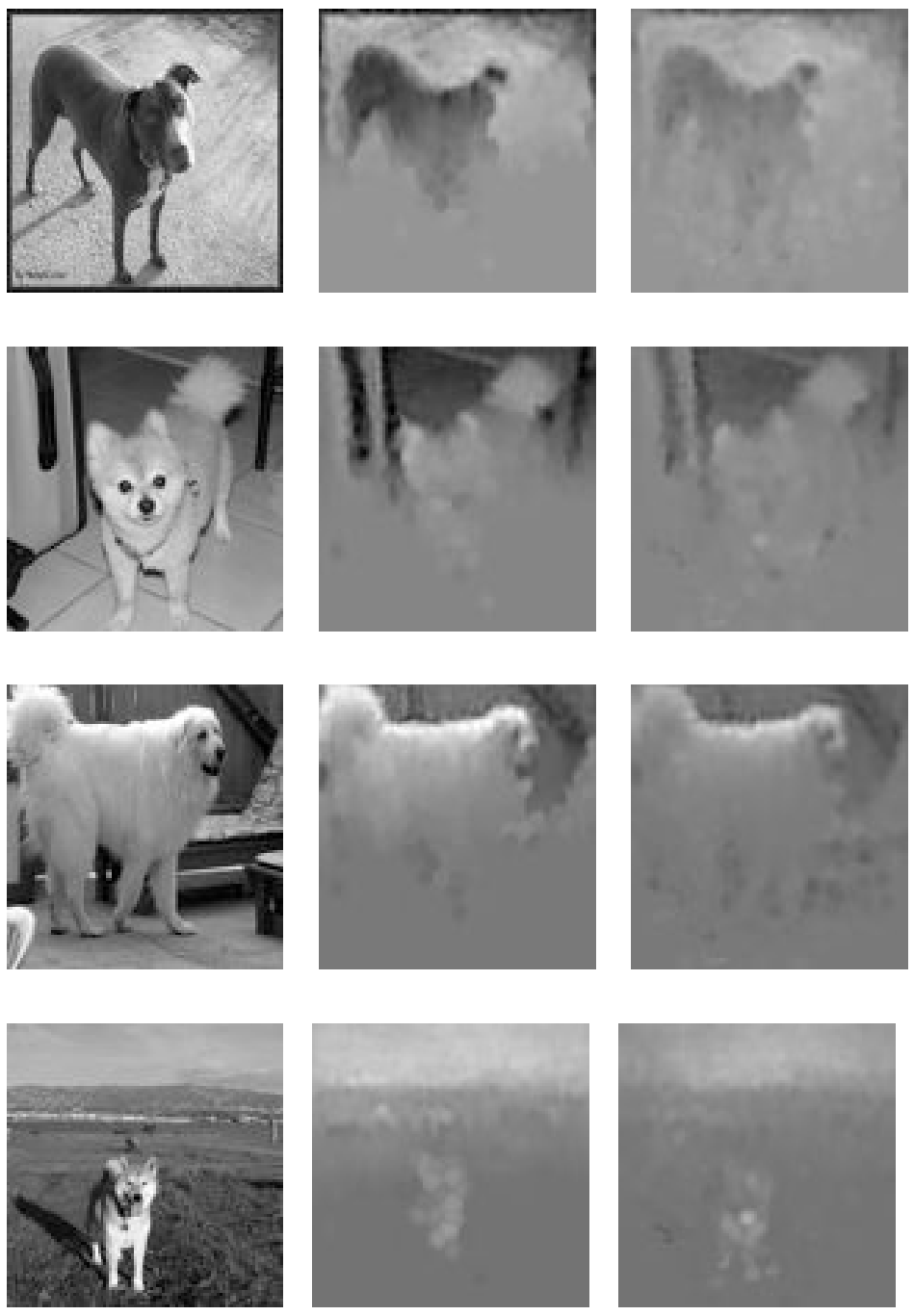

Figure 17: Reconstructed animal images of size 100x100 (test data), with 500 *sparse dictionary elements (non-sparse codes)*. In each row, the original image is on the left, and the reconstructions, computed with ODL and NODL (our algorithm), are in the center and right respectively.

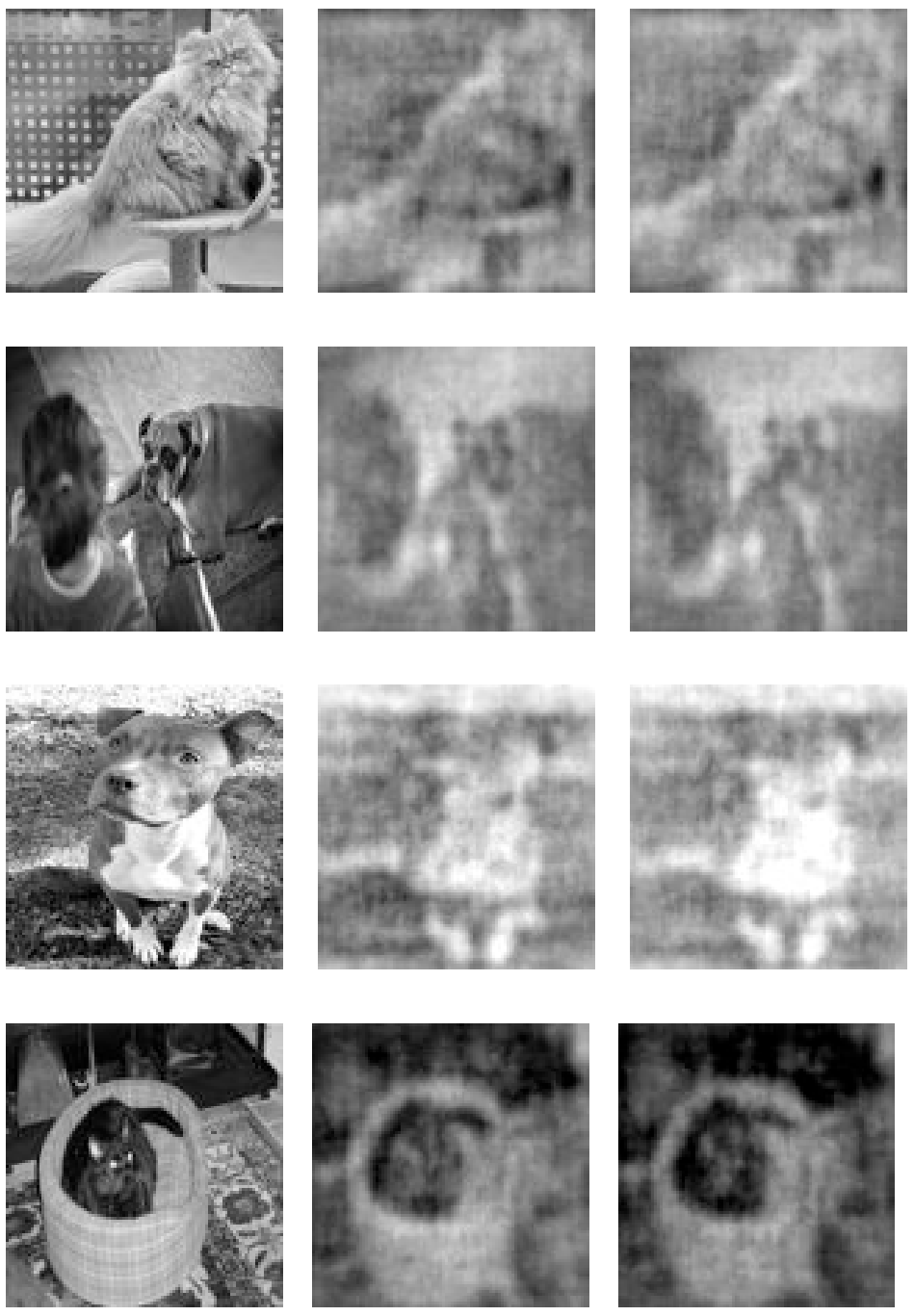

Figure 18: Reconstructed animal images of size 100x100 (test data), with 500 *non-sparse dictionary elements (non-sparse codes)*. In each row, the original image is on the left, and the reconstructions, computed with ODL and NODL (our algorithm), are in the center and right respectively.

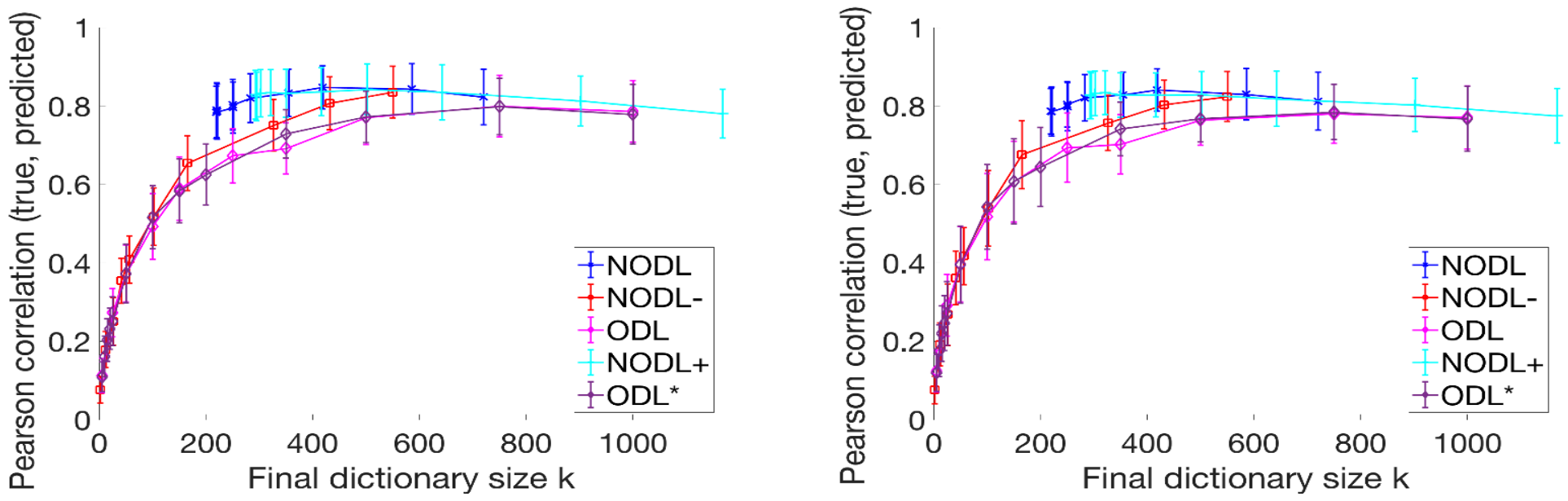

(a) 2nd domain (Flowers) (b) 2nd domain (Animals)

Figure 19: Extension of Fig. 2, with the results for the ODL* version of ODL where occasional "dead" elements are reinitialized with random values.

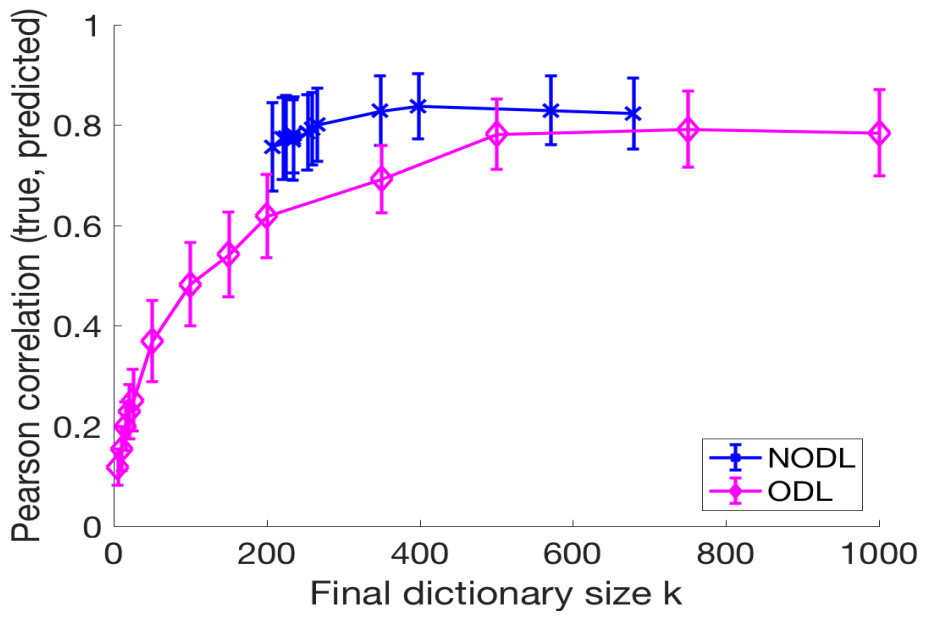

(a) Training Order: 1st domain (Oxford), 2nd domain (Flowers)

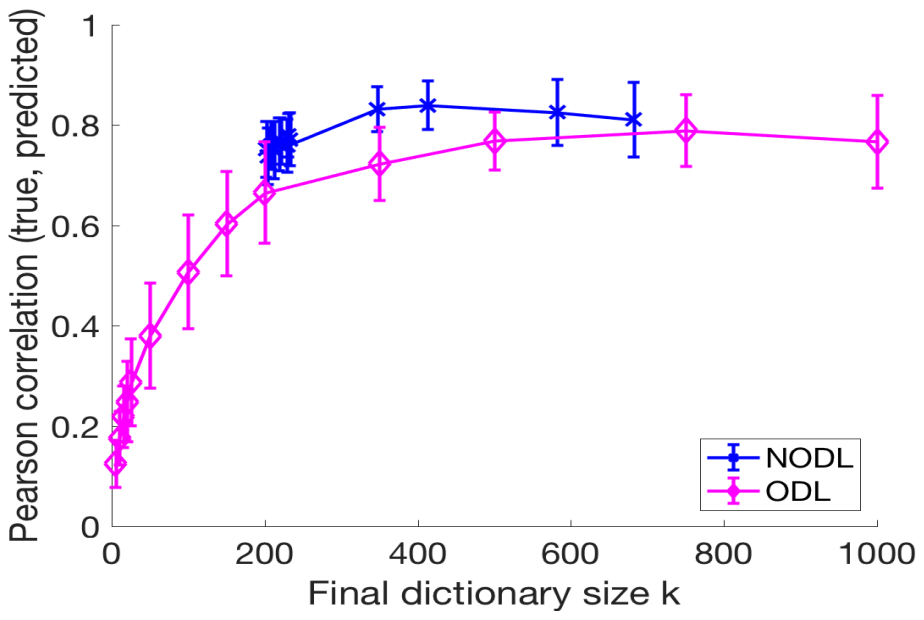

(b) Training Order: 1st domain (Oxford), 2nd domain (Flowers)

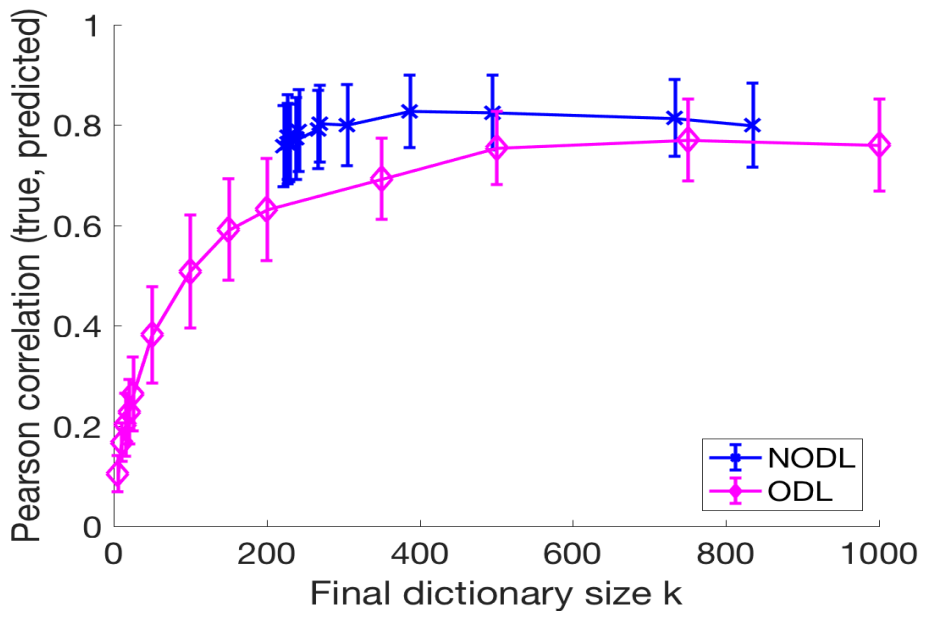

(c) Training Order: 1st domain (Flowers), 2nd domain (Oxford)

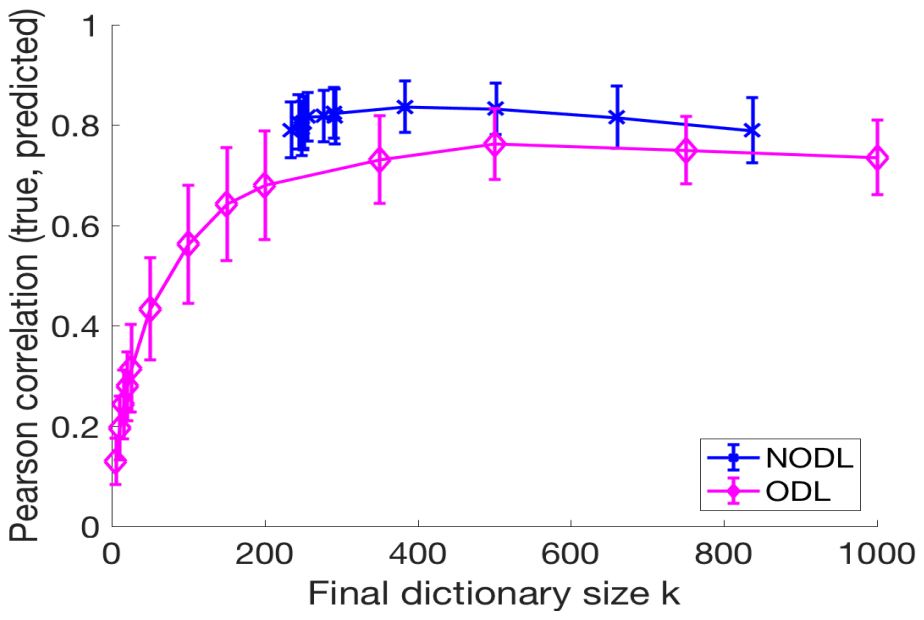

(d) Training Order: 1st domain (Flowers), 2nd domain (Animals)

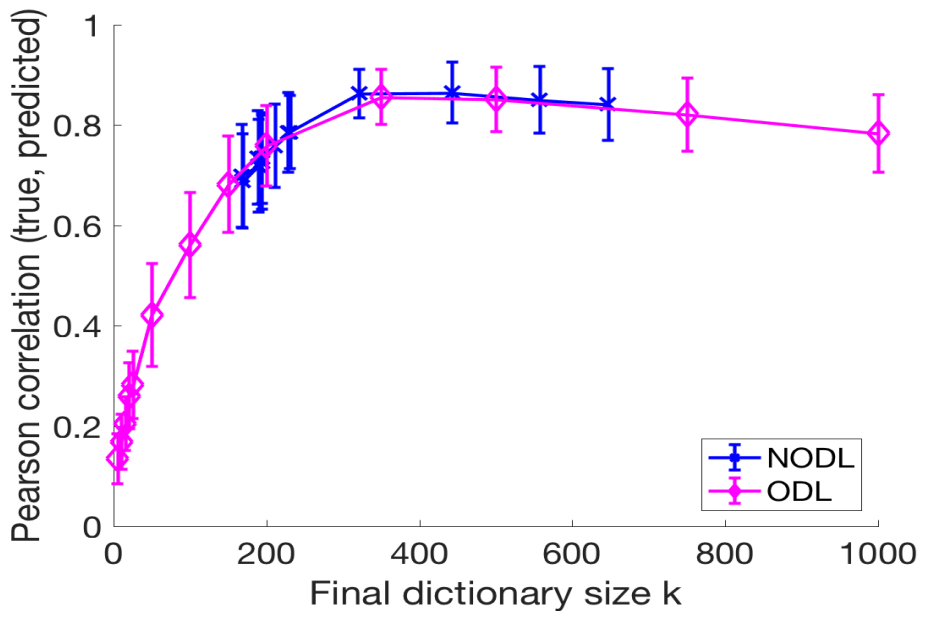

(e) Training Order: 1st domain (Animals), 2nd domain (Oxford)

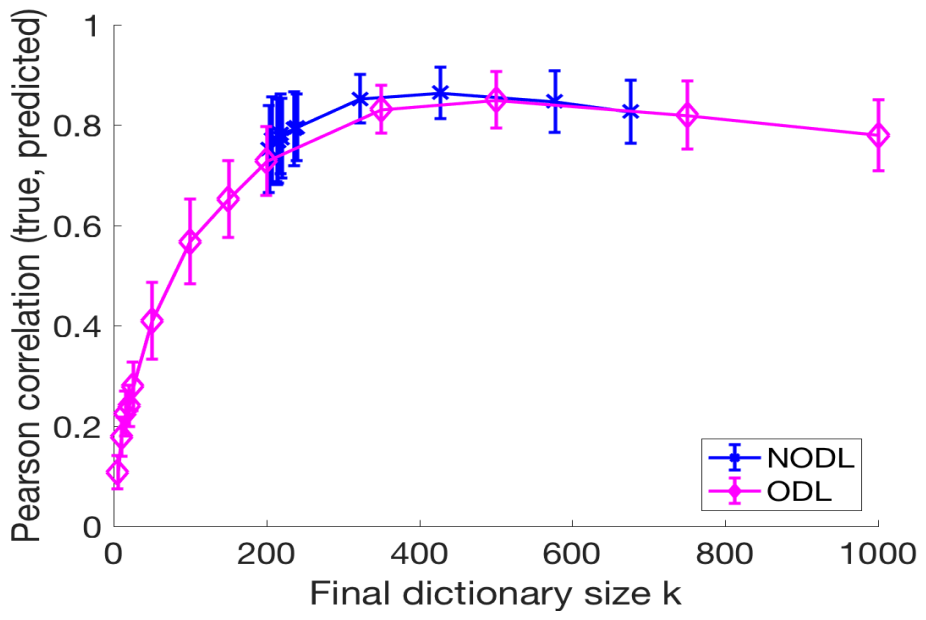

(f) Training Order: 1st domain (Animals), 2nd domain (Flowers)

Figure 20: Evaluating the effects of the input data order; the experimental setup coincides with the one used to produce Fig. 2 (32x32 images). Different processing orders of the available datasets are used during the training phase; performance results on the test subset taken from the second domain are presented.

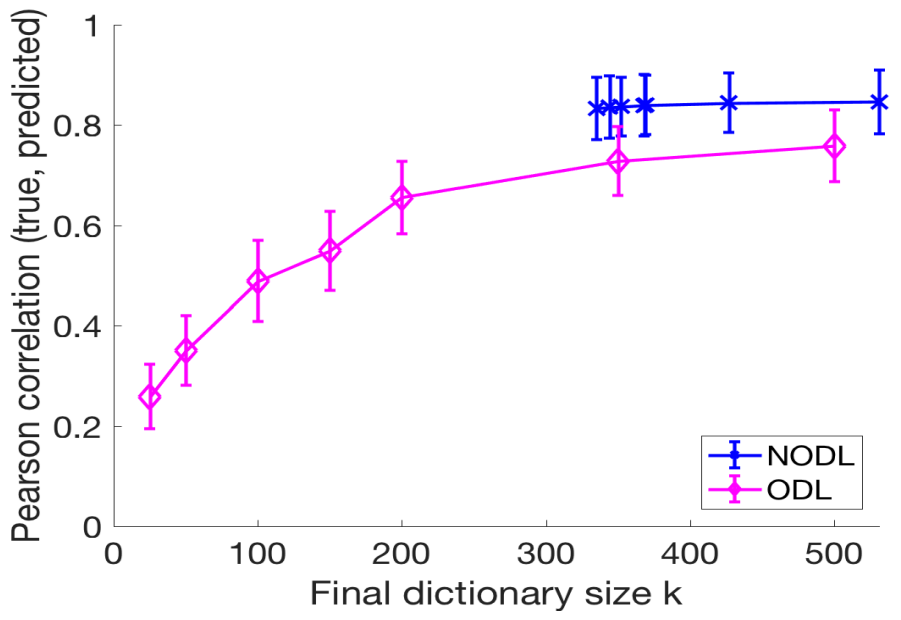

(a) Batch Size- 125

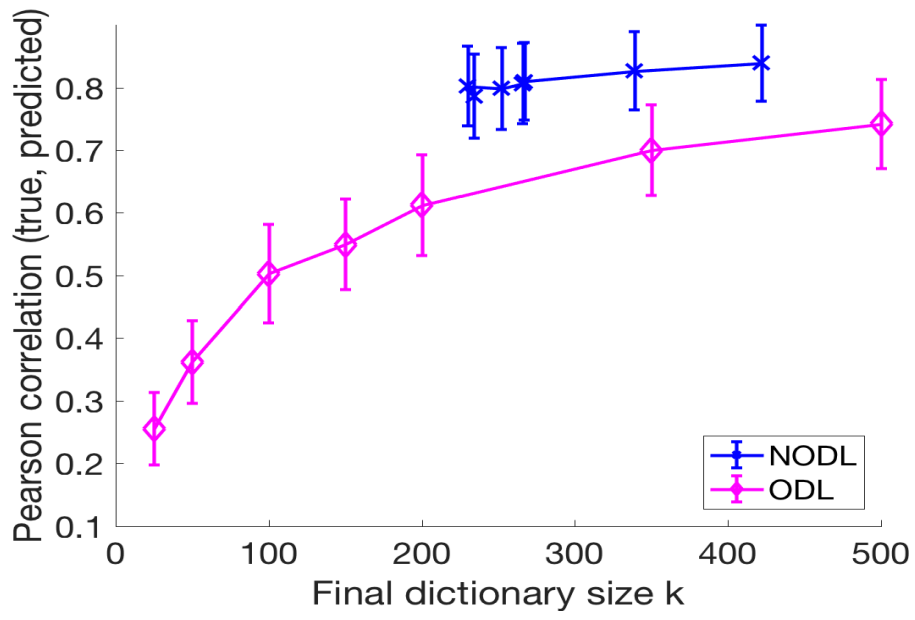

(b) Batch Size- 200

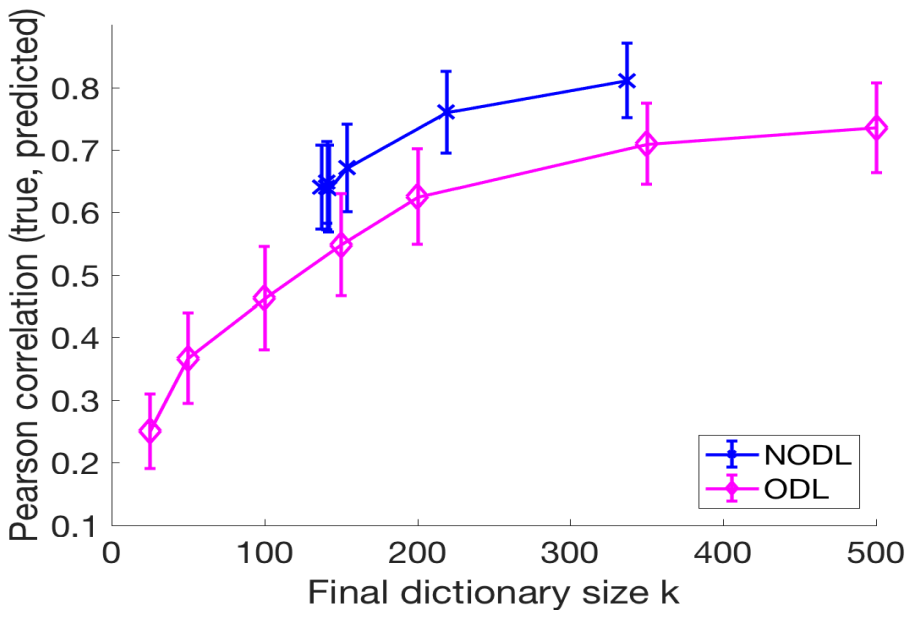

(c) Batch Size- 350

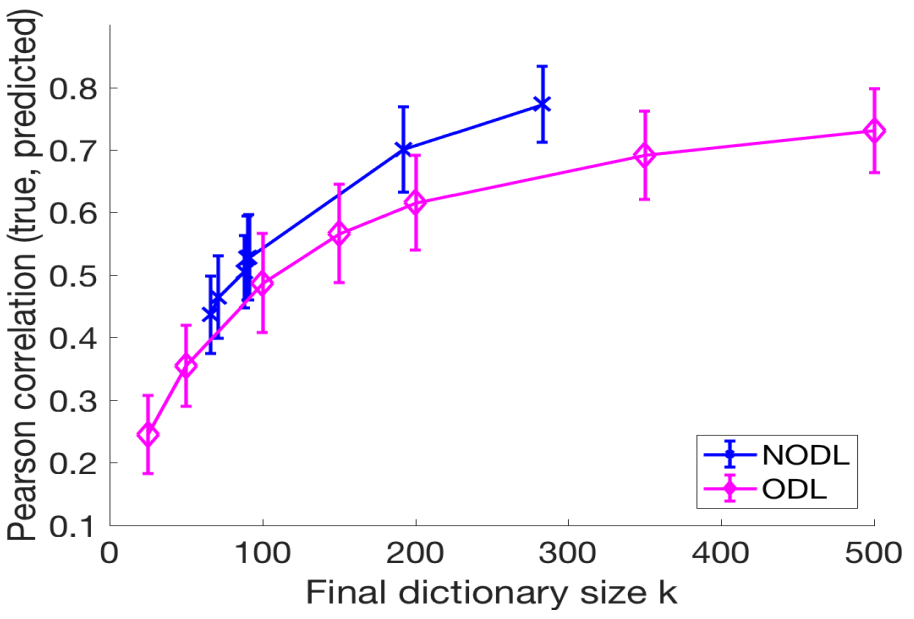

(d) Batch Size- 500

Figure 21: Evaluating the effects w.r.t. batch size while keeping the other experimental settings same as the ones used to produce Fig. 2 (32x32 images).

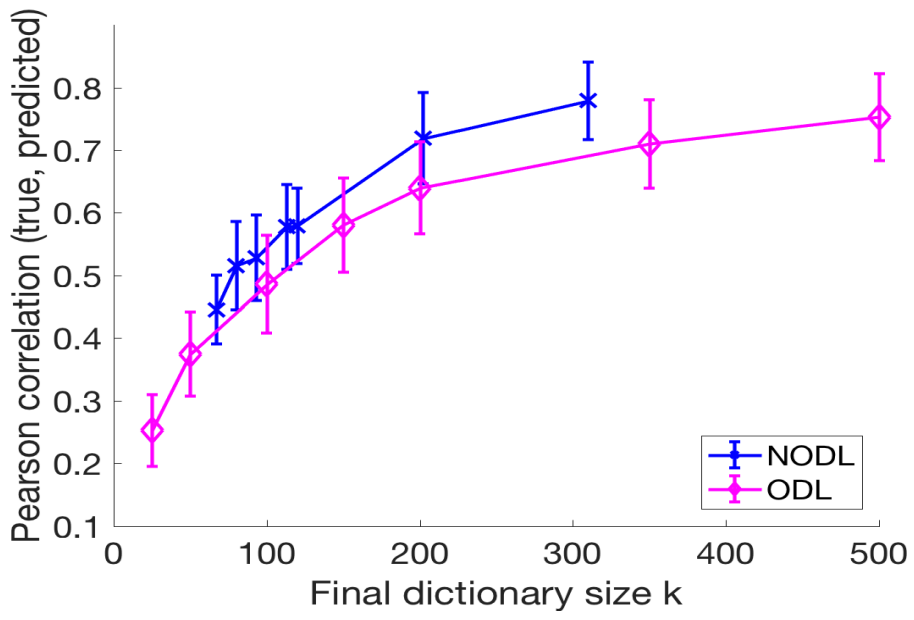

(a) $c_k = 10$

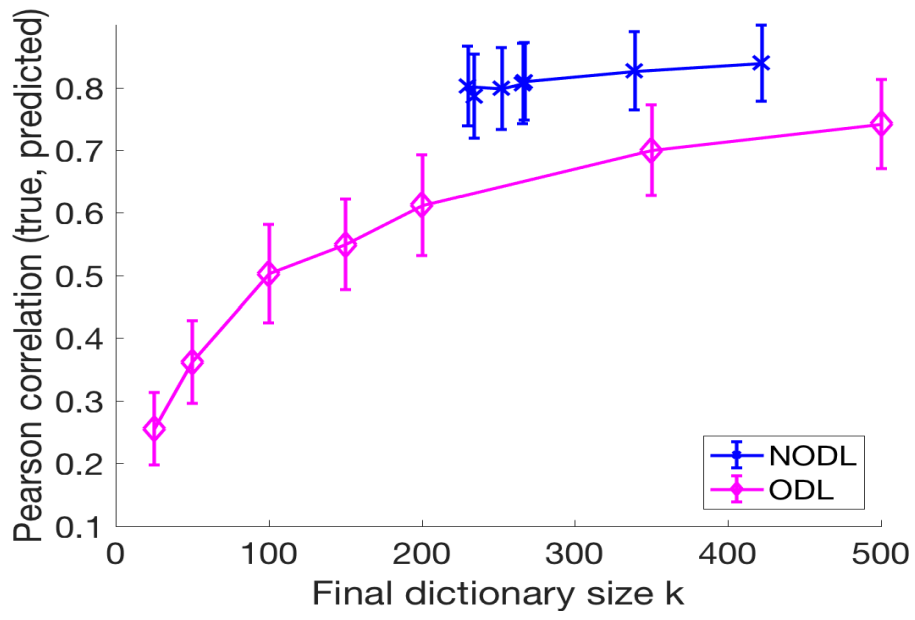

(b) $c_k = 50$

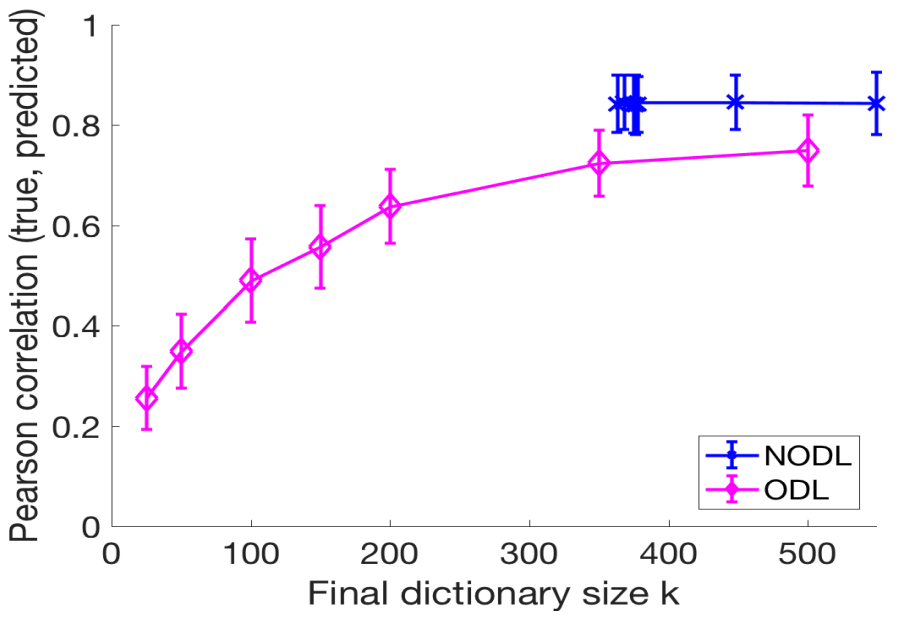

(c) $c_k = 100$

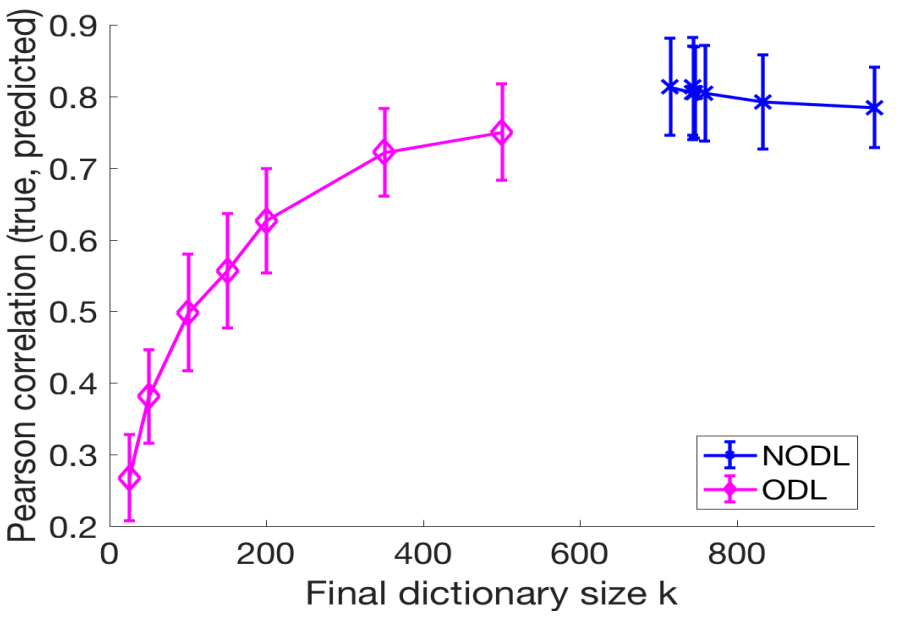

(d) $c_k = 250$

Figure 22: Evaluating the effects w.r.t. the tuning parameter $c_k$ (the upper bound on the number of new elements added in a batch) while keeping the other experimental settings same as the ones used to produce Fig. 2 (32x32 images).

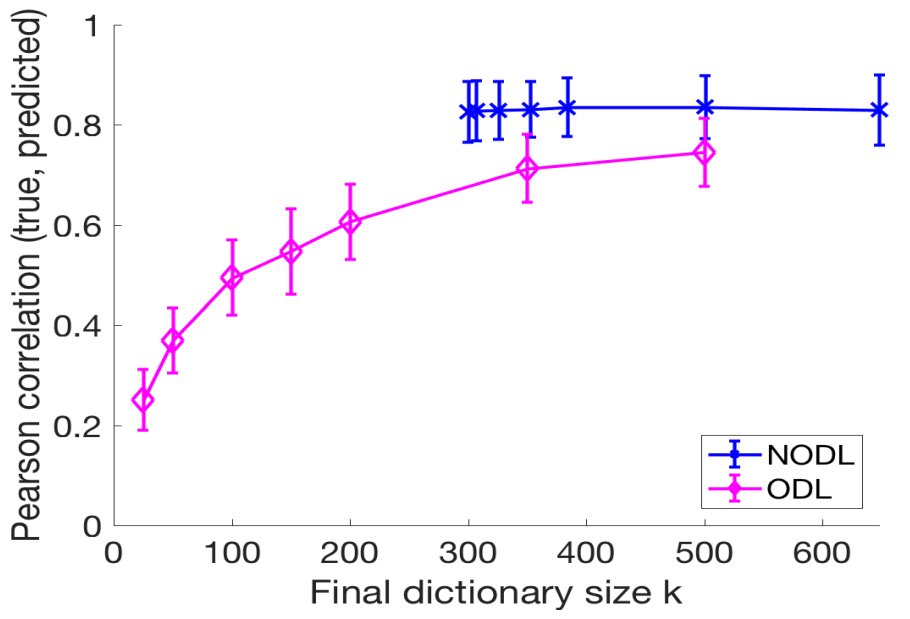

(a) $\lambda_g = 3e-3$

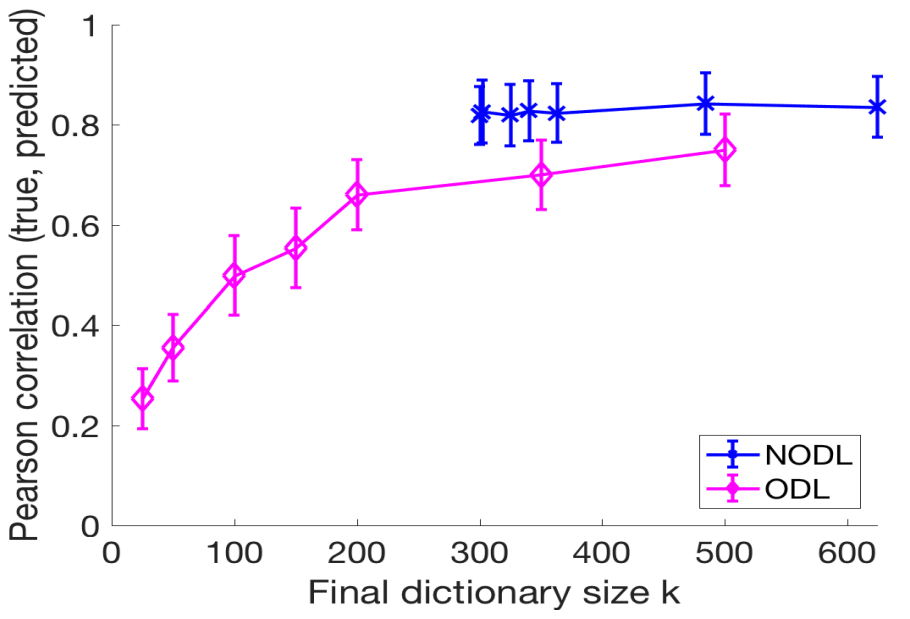

(b) $\lambda_g = 1e-2$

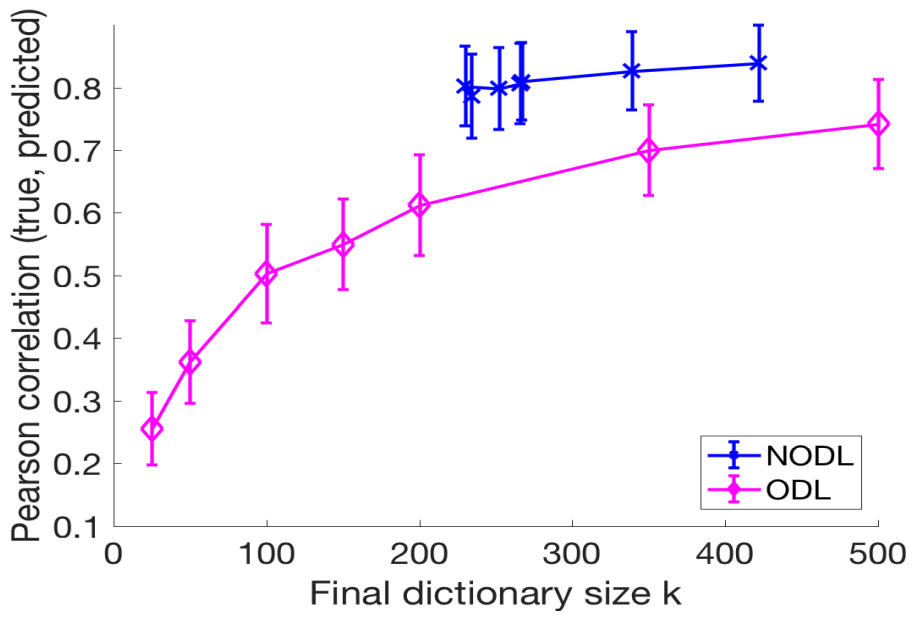

(c) $\lambda_g = 3e-2$

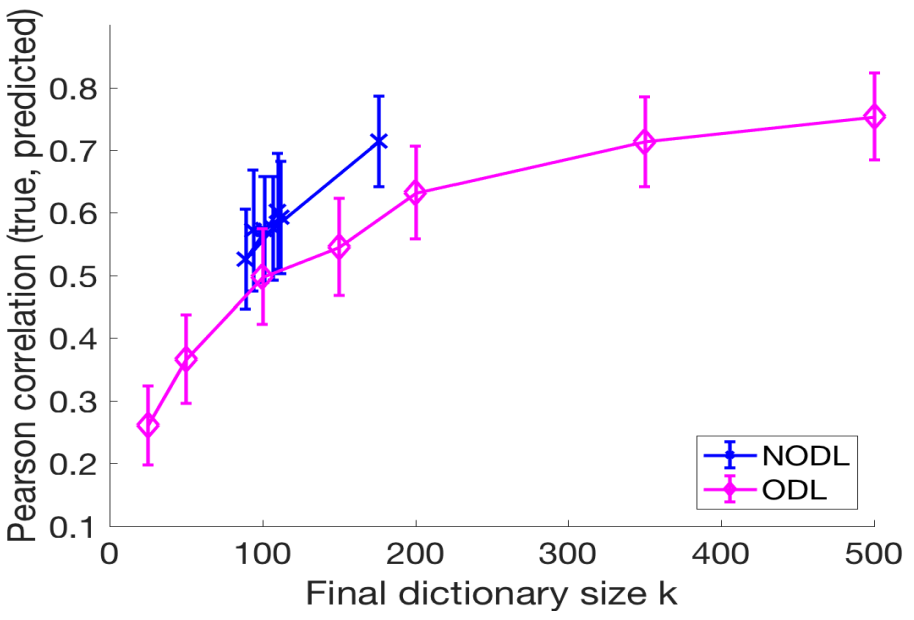

(d) $\lambda_g = 5e-2$

Figure 23: Evaluating the effects w.r.t. the tuning parameter $\lambda_g$ (the regularization parameter for the killing of "weak" elements) while keeping the other experimental settings same as the ones used to produce Fig. 2 (32x32 images).

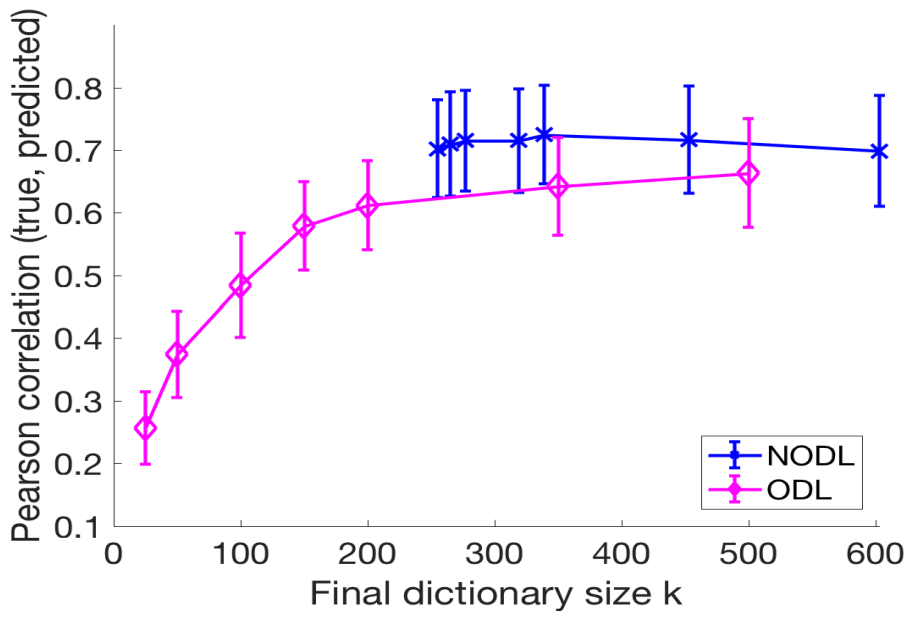

(a) $\beta_c = 100$

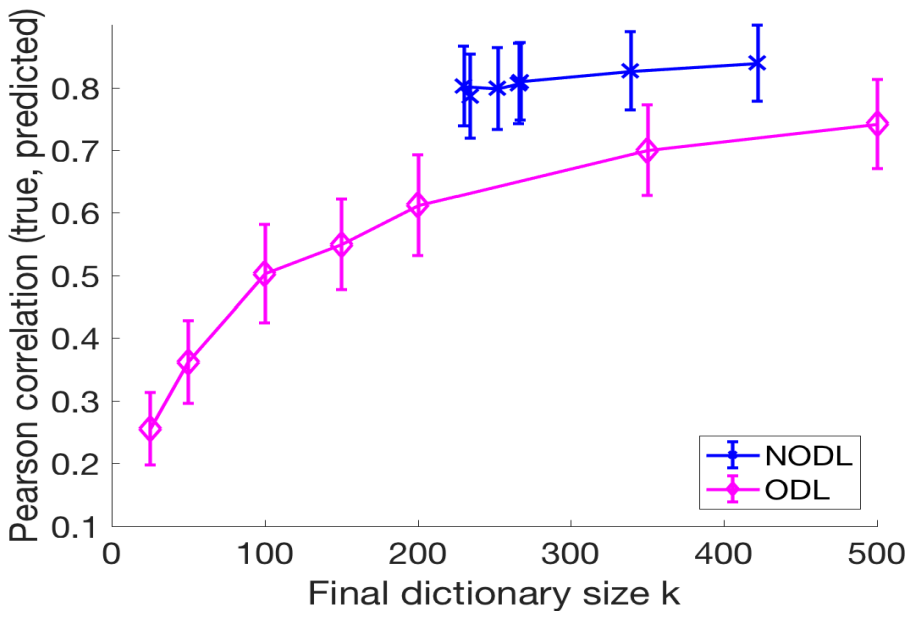

(b) $\beta_c = 200$

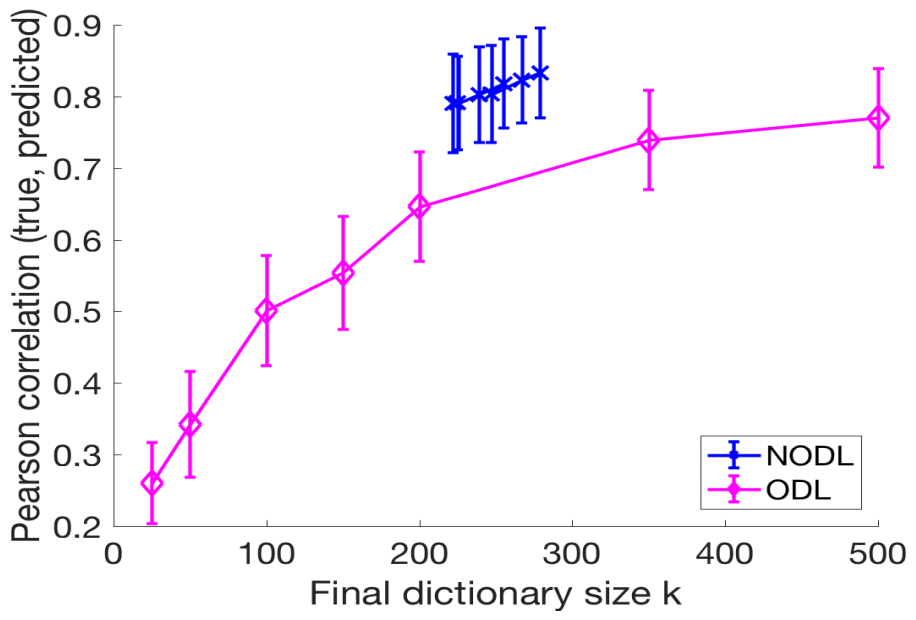

(c) $\beta_c = 500$

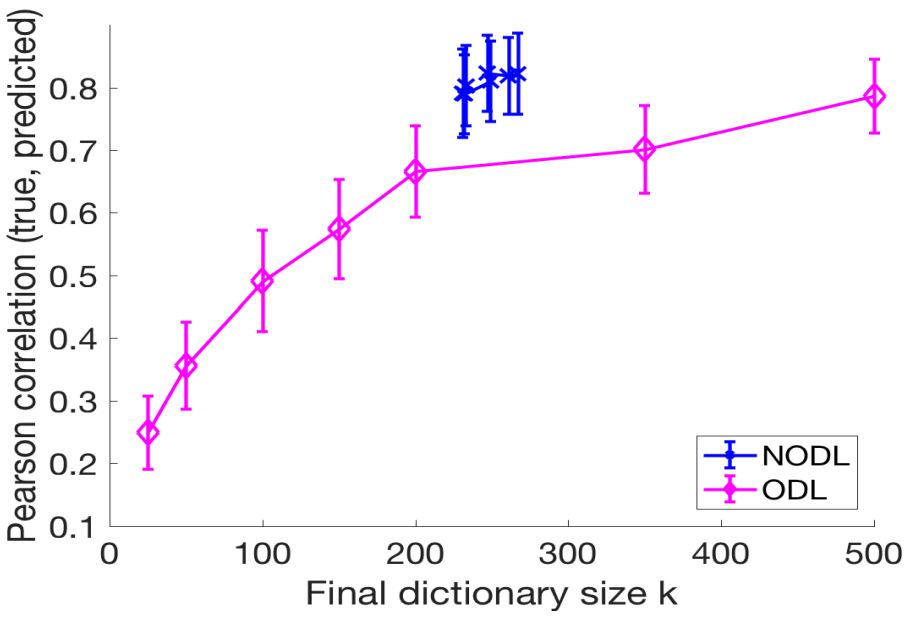

(d) $\beta_c = 1000$

Figure 24: Evaluating the effects w.r.t. the tuning parameter $\beta_c$ (the number of non-zeros in a code) while keeping the other experimental settings same as the ones used to produce Fig. 2 (32x32 images).

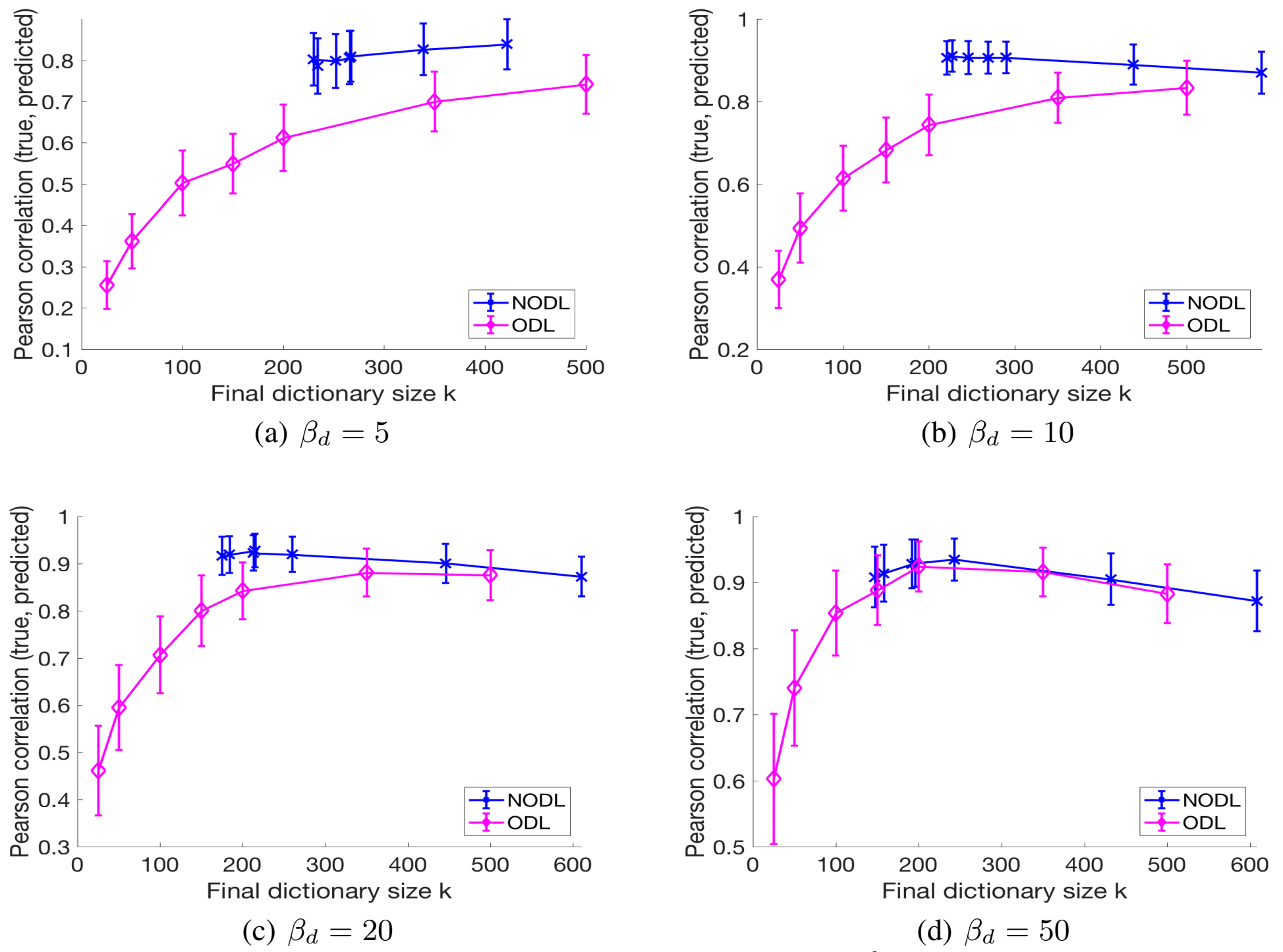

Figure 25: Evaluating the effects w.r.t. the tuning parameter $\beta_d$ (the number of non-zeros in a dictionary element) while keeping the other experimental settings same as the ones used to produce Fig. 2 (32x32 images).

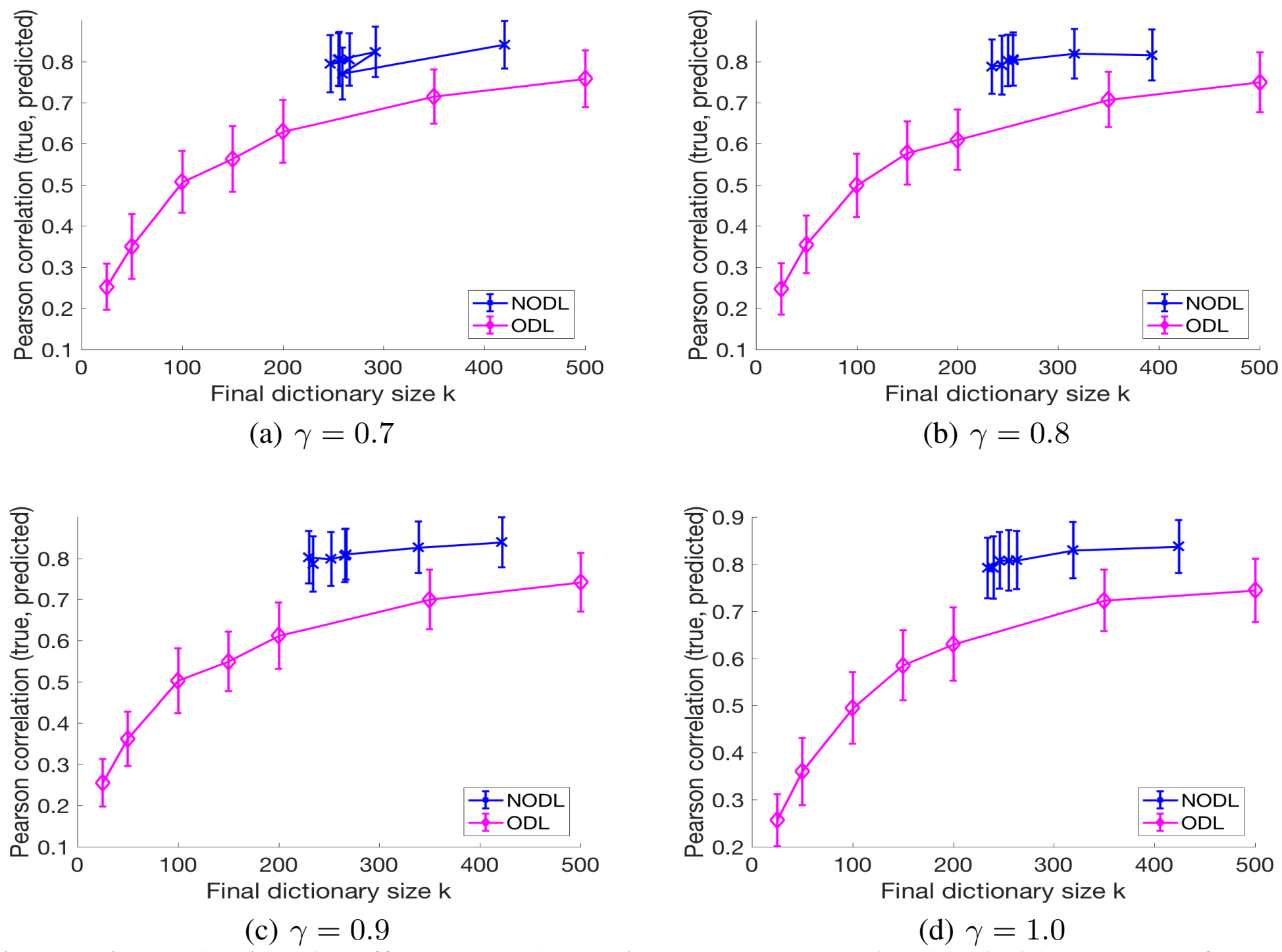

Figure 26: Evaluating the effects w.r.t. the tuning parameter $\gamma$ (the threshold parameter for conditional neurogenesis) while keeping the other experimental settings same as the ones used to produce Fig. 2 (32x32 images).